\pdfoutput=1

\documentclass[11pt]{article}

\usepackage{authblk}
\usepackage[]{ACL2023}

\usepackage{multirow}
\usepackage{times}
\usepackage{latexsym}
\usepackage[normalem]{ulem}
\useunder{\uline}{\ul}{}
\usepackage{amsmath}
\usepackage{amssymb}
\usepackage{latexsym}
\usepackage{algorithm,algorithmic}
\usepackage{babel}
\usepackage{graphicx}
\usepackage[T1]{fontenc}

\newcommand{\method}{\texttt{Self-AMPLIFY}}

\usepackage[utf8]{inputenc}

\usepackage{microtype}

\usepackage{inconsolata}

\title{Self-AMPLIFY : Improving Small Language Models\\ with Self Post Hoc Explanations}

\author[1,2]{Milan Bhan}
\author[1]{Jean-Noël Vittaut}
\author[2]{Nicolas Chesneau}
\author[1]{Marie-Jeanne Lesot}
\affil[1]{Sorbonne Université -- CNRS -- LIP6, Paris, France}
\affil[2]{Ekimetrics, Paris, France}
\affil[ ]{\texttt{\{milan.bhan, nicolas.chesneau\}@ekimetrics.com}}
\affil[ ]{\texttt{\{jean-noel.vittaut, marie-jeanne.lesot\}@lip6.fr}}

\begin{document}
\maketitle
\begin{abstract}

Incorporating natural language rationales in the prompt and In-Context Learning (ICL) have led to a significant improvement of Large Language Models (LLMs) performance. However, generating high-quality rationales require human-annotation or the use of auxiliary proxy models. In this work, we propose \method\ to automatically generate rationales from post hoc explanation methods applied to Small Language Models (SLMs) to improve their own performance. \method\ is a 3-step method that targets samples, generates rationales and builds a final prompt to leverage ICL. \method\ performance is evaluated on four SLMs and five datasets requiring strong reasoning abilities. \method\ achieves good results against competitors, leading to strong accuracy improvement. \method\ is the first method to apply post hoc explanation methods to autoregressive language models to generate rationales to improve their own performance in a fully automated manner. 
\end{abstract}

\section{Introduction} 
Autoregressive Large Language Models (LLMs) such as GPT-3~\cite{brown_language_2020}, PaLM~\cite{chowdhery2023palm} or LaMDA~\cite{thoppilan2022lamda}, have made significant advancements in a wide range of NLP tasks. These models have demonstrated so-called "emergent abilities"~\cite{ schaeffer_are_2023}, including in-context learning (ICL), instruction following and reasoning~\cite{wei2022emergent}. ICL (see~\citet{dong_survey_2023} for a recent survey) involves learning from a few examples integrated into the prompt without fine tuning the model.

\label{intro}
\begin{figure}[t]{\centering}
\begin{center}
\includegraphics[scale=0.6]{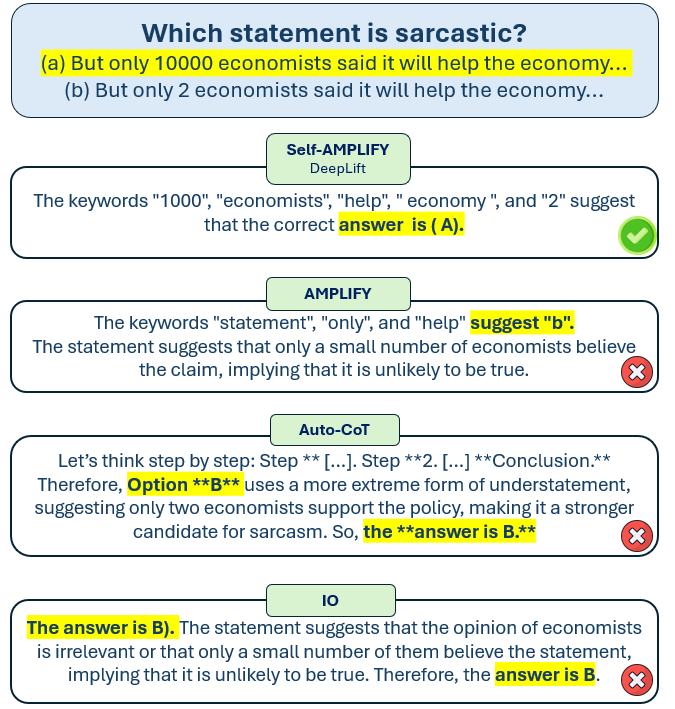}
\caption{Example of four responses to a question from the Snarks dataset, generated from different ICL prompting strategies. Traditional input-output (\texttt{IO}) prompting, \texttt{Auto-CoT}~\cite{zhang_automatic_2022} and \texttt{AMPLIFY}~\cite{krishna_post_2023} fail to answer properly, whereas \method\ generates important tokens as a rationale before correctly answering.}
\label{fig:intro_fig}
\end{center}
\end{figure}

LLMs' emergent abilities have been leveraged to enhance performance by incorporating human-annotated intermediate reasoning steps within the context, called \emph{rationales}~\cite{cot}. By learning to sequentially generate (1) the reasoning steps through rationales and (2) the final answer, LLMs have reached state-of-the-art performance in complex tasks requiring reasoning abilities such as commonsense or symbolic reasoning. To overcome the need for human annotation, automatic rationale generation methods have been proposed. \texttt{AMPLIFY}~\cite{krishna_post_2023} has demonstrated that rationales can be  generated from smaller proxy supervised Language Models (LMs) to enrich the prompt to enhance the performance of LLMs. \texttt{AMPLIFY} targets promising instances to be integrated into the final prompt using the proxy model and automatically builds rationales based on post hoc attribution explanation methods~\cite{molnar_interpretable_2020} applied to this proxy model.

Recently, small autoregressive LMs (SLMs), with fewer than 14 billion parameters, have emerged, such as Mistral~\cite{jiang2023mistral}, Zephyr~\cite{tunstall_zephyr_2023} or Gemma~\cite{gemma}. They achieve performance sometimes approaching that of LLMs' on common benchmarks: their smaller size makes them computationally efficient while maintaining a high level of accuracy. In particular, classical post hoc attribution methods such as \texttt{KernelSHAP}~\cite{lundberg_unified_2017} or \texttt{DeepLift}~\cite{shrikumar_learning_2017} become affordable to explain SLMs' prediction, despite their high computational cost of these methods.
 
In this paper, we propose \method, an extension of the \texttt{AMPLIFY} framework for SLMs that does not need an auxiliary model nor human annotations. The main contributions of the \method\ framework are as follows: (i) promising instances to be integrated into the final prompt are targeted only using the considered SLM's prediction, (ii) post hoc explanation methods are applied to the SLM itself to automatically generate rationales as a self-improving signal, (iii) three types of post hoc explanations methods are implemented: post hoc attributions, self \texttt{topk} explanations and self free text rationales. 

As an illustration, Figure~\ref{fig:intro_fig} shows three responses to a question from the Snarks~\cite{beyond_imitation_game} dataset respectively obtained using  the proposed \method, a classical prompting approach, \texttt{IO}, a rationale enhanced approach, \texttt{Auto-CoT}~\cite{zhang_automatic_2022} and \texttt{AMPLIFY}. \method\ succeeds to generate the good answer whereas its three competitors fail.

Experimental results  discussed in Section~\ref{xp} show that \method\ leads to a performance gain on a wide range of datasets as compared to \texttt{IO}, \texttt{Auto-CoT} and \texttt{AMPLIFY}. As a result, we show that post hoc explanation methods of various kinds can be directly applied to the SLM to generate automatically rationales to self-improve. Unlike the original \texttt{AMPLIFY} framework, proxy fine tuned models are no longer needed to increase LMs' performance, making \method\ more autonomous and flexible.

\section{Background and Related Work}
\label{sec:bk_rw}

In this work, we consider in-context learning (ICL), where a few samples are provided to an autoregressive LM within the prompt to perform a particular NLP task. In this section we recall some basic principles of post hoc explanations and existing methods that generate rationales to enhance LMs' performance by enriching prompts.

\subsection{Post Hoc Explanations Background}
\paragraph{Attribution method.} Attribution methods compute an importance score for each input feature to explain the model output. Two types of methods can be distinguished: \textit{perturbation-based} and \textit{gradient-based}~\cite{zhao_explainability_2023}. 

The former perturbs and resamples feature values to compute feature importance. Two common examples are LIME~\cite{ribeiro_why_2016} and KernelSHAP~\cite{lundberg_unified_2017}. However, these methods are computationally expensive due to the numerous inferences required. 

On the other hand, gradient-based approaches estimate feature importance through the model backpropagated gradient activity. Two common examples are Integrated Gradients~\cite{sundararajan_axiomatic_2017} and DeepLift~\cite{shrikumar_learning_2017}. However, these methods are computationally expensive due to the need to compute gradients. Therefore, to the best of our knowledge, they have not been yet applied to autoregressive LLMs.  

\paragraph{Post hoc free text self-rationales.} Free text rationales are natural language intermediate reasoning steps that justify a model's prediction (see~\citet{gurrapu_rationalization_2023} for a recent survey) or favor reasoning in LLMs~\cite{huang_towards_2023}. Post hoc self-rationale generation involves directly prompting LM's to explain their prediction in free text given their answer~\cite{huang_can_2023, self_consistency_madsen}. Post-hoc self-rationales contrast with attribution numerical vector explanations in terms of their higher level of abstraction.

\begin{figure*}[t]{\centering}
\begin{center}
\includegraphics[scale=0.70]{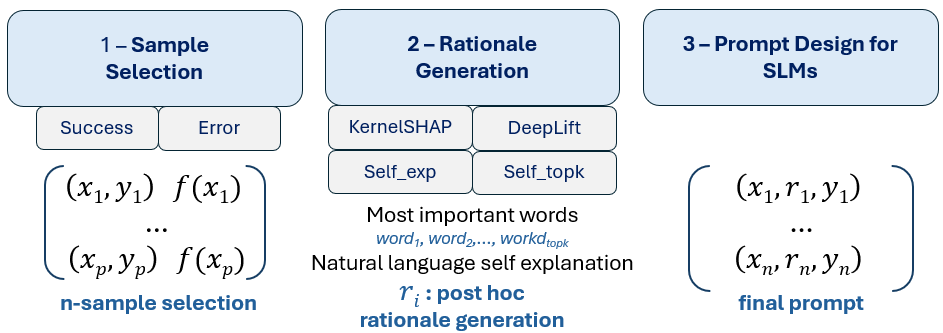}
\caption{\method\ overview. \method\ is a 3-step approach generating rationales to self-improve a SLM in a ICL setting. (1) Promising samples are targeted following two selection strategies: \texttt{success} or \texttt{error}. (2) Rationales are generated based on a post hoc explanation method: \texttt{KernelShap}, \texttt{DeepLift}, \texttt{Ph-CoT} or \texttt{Self\_topk}. (3) The final ICL prompt is built based on the previously generated rationales.}  
\label{fig:global_flow}
\end{center}
\end{figure*}

\subsection{Related Work}
\label{related_work}
This section introduces two categories of methods for generating rationales aimed at enriching the prompt and encouraging LLMs to engage in reasoning rather than merely providing answers.
\paragraph{Human-annotated rationales.}
Firstly, rationales can be generated manually. Several handcrafted benchmarks have been proposed to either train language models to generate rationales or to assess language models' ability to generate rationales aligned with human annotations, such as e-SNLI~\cite{esnli} or ERASER~\cite{eraser}. Chain-of-Thought (CoT)~\cite{cot} adds human-annotated rationale steps to the standard ICL prompt template ($x$, $y$) to construct a \textit{explain-then-predict} template ($x$, $r$, $y$) where $x$ is the input text, $y$ is the expected answer and $r$ is the provided rationale. CoT extensions have been proposed to aggregate multiple reasoning paths~\cite{wang_self-consistency_2023} or to enable LLMs to explore multiple promising reasoning paths~\cite{yao_tree_2023} during text generation. These approaches significantly improve LLMs' performance on NLP tasks requiring reasoning capabilities. 
Another way of using rationales to enrich the ICL prompt consists in appending the rationale after the answer in a \textit{predict-then-explain} manner, as ({$x, y, r$}), resulting in a relative performance gain~\cite{lampinen_can_2022} as compared to the (${x, r, y}$) design. 

However relying on human-annotated rationales makes these methods costly and not automatable. Moreover, they require strong reasoning capabilities and often yield significant performance gains only for LLMs larger than a certain size~\cite{cot}.

\paragraph{Automatically generated rationales.}
Automatic rationale generation eliminates the need for human-annotated rationales. Automatic Chain-of-Thought prompting (\texttt{Auto-CoT})~\cite{zhang_automatic_2022} proposes to generate automatically natural language rationales by prompting the LLM to "think step by step". A Sentence-Transformer~\cite{reimers_sentence-bert_2019} is used to cluster input texts in order to generate one CoT rationale per cluster, making the approach dependent on this auxiliary Sentence Transformer. Then, the LLM's prediction~$\hat{y}$ is integrated to construct the final prompt ($x$, $r$, $\hat{y}$). However, \texttt{Auto-CoT} is prone to include incorrect demonstrations and low-quality samples in the prompt, since it does not take the ground truth answer for the final prompt construction.

\texttt{AMPLIFY}~\cite{krishna_post_2023} automatically generates rationales from post hoc numeric attribution methods from an auxiliary fine tuned proxy model. The latter is initially fine tuned on a corpus of interest to generate relevant explanations. Then, a $n$-shot sample selection is performed using the same proxy model to identify misclassified instances. These samples are then added to the ICL prompt, following a ($x$, $r$, $y$) template. Therefore, \texttt{AMPLIFY} relies heavily on the use of the auxiliary proxy model, both at the $n$-shot targeting and the rationale generation steps. While \texttt{AMPLIFY} yields significant performance gain as compared to classical prompting, it has only been tested on GPT-3 and GPT-3.5. Moreover, \texttt{AMPLIFY} does not incorporate free text rationales in its framework.

\section{Proposed approach: \method} \label{approach}

This section describes the architecture of \method, an extension of the \texttt{AMPLIFY}~\cite{krishna_post_2023} framework. As sketched in Figure~\ref{fig:global_flow} and detailed in the next subsections, this framework enriches prompts with self-generated rationales in a fully automated manner to enhance SLMs' performance in ICL settings. By generating rationales directly from the SLM, \method\ differs from \texttt{AMPLIFY} in that it does not depend on any auxiliary fine-tuned proxy model and the data used to train it. Therefore, post-hoc explanation methods are leveraged to self-improve SLM fully automatically.  

\subsection{\method\ overview}
As shown in Figure~\ref{fig:global_flow} and detailed in the following, \method\ is a 3-step  approach that takes as input an autoregressive SLM~$f$ and a corpus of texts $\mathcal{T}$ from which the $n$-shot sample is generated. Each input text is associated with an expected answer, belonging to a label space denoted~$\mathcal{L}$.

\paragraph{(i) $n$-shot Sample Selection.} This step aims to select input texts that will be added to the final prompt. \method\ employs two simple yet efficient selecting strategies only based solely on $f$ prediction, eliminating the need of an auxiliary model as in the \texttt{AMPLIFY} framework. 


\paragraph{(ii) Rationale Generation.} Rationales are generated for the previously selected texts by applying post hoc explanation methods to $f$ itself. This way, unlike \texttt{AMPLIFY}, rationales are not generated from a fine tuned side proxy model. We implements 3 types of post-hoc explanation methods to  generate rationales directly from $f$, making \method\ more versatile.

\paragraph{(iii) Prompt Design for SLMs.} The final prompt is constructed based on the previously generated rationales. Each generated rationale is added between its related input text and ground truth answer. The enriched sample is finally used to make the prediction on the test set.  

\subsection{$n$-shot Sample Selection}
\label{sub:selec}
The first step involves selecting $n$ instances from the text corpus $\mathcal{T}$ for inclusion in the final prompt. 

\method\ employs two selection strategies based solely on $f$ prediction: \texttt{success} and \texttt{error}. The \texttt{success} strategy selects text instances correctly predicted by $f$ in a standard prompt setting, whereas the \texttt{error} strategy selects ones incorrectly predicted. To determine if an instance of interest $x \in \mathcal{T}$ is correctly predicted, we append the text "The answer is" to the initial prompt to guide $f$ next token prediction. Therefore, the next token is more likely to be predicted in the correct format as in~\citet{kojima2022large}, i.e with the next token predicted in the label space $\mathcal{L}$. Denoting $y$ the ground truth, the model prediction is categorized as a success if $f(x) = y$ and an error if $f(x)\neq y$ with $f(x) \in \mathcal{L}$. Otherwise, $x$ is disgarded. 

The \texttt{success} strategy relies on the idea that "the higher the prediction certainty, the more relevant the explanation"~\cite{bhan_evaluating_2023}. Conversely, the \texttt{error} strategy relies on the idea that adding  misclassified examples may avoid similar misclassifications on the test set. We assess the impact of the selection strategy on $f$ performance in Section~\ref{xp}. This way, regardless of the selection strategy, \method\ does not rely on a proxy additional model to select samples, making it more flexible than other methods.

\begin{figure}[t]{\centering}
\begin{center}
\includegraphics[scale=0.75]{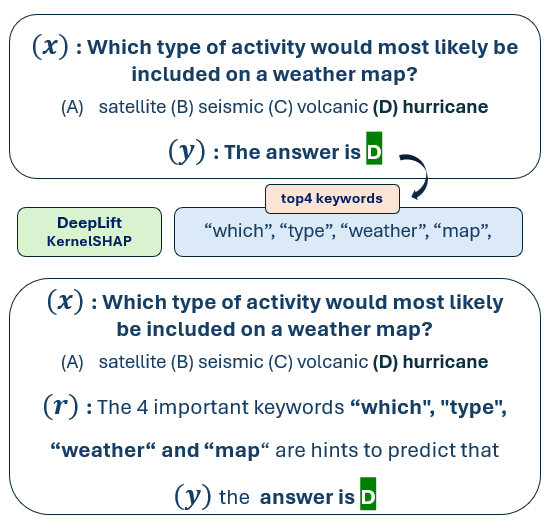}
\caption{\method\ rationale generation step with a post hoc attribution method. Here,  \texttt{DeepLift} or \texttt{KernelShap} is applied to the SLM to explain the answer D. The 4 most important tokens are targeted and the final rationale $r$ is constructed based on these keywords. The ($x$, $r$, $y$) triplet is finally added to the ICL prompt.}
\label{fig:ratio_generation}
\end{center}
\end{figure}

\subsection{Rationale Generation}
The rationale generation step is summarized in Figure~\ref{fig:ratio_generation}. Once the $n$-shot sample is created, rationales are generated by computing post hoc explanation from $f$ directly. \method\ differs from \texttt{AMPLIFY} in that it generates rationales without the use of an auxiliary fine tuned model. In addition, \method\ implements 3 types of post hoc explanations to generate natural language rationale: post hoc attributions (\texttt{DeepLift} and \texttt{KernelSHAP}), post hoc \texttt{Self\_topk} explanations and post hoc CoT (\texttt{Ph-CoT}) rationales where \texttt{AMPLIFY} only implements attribution methods, making \method\ more versatile. Post-hoc explanations are computed to explain each $(x, y)$ pair to finally build their associated rationales~$r$. 

\texttt{DeepLift} and \texttt{KernelShap} are computed to explain the $(x, y)$ pair, i.e. $f$ output neuron related to $y$. \texttt{DeepLift} decomposes the neural network prediction by backpropagating the contributions of all neurons in the network to each input feature. Attribution scores are computed with respect to a chosen baseline. We define this baseline so that attribution is only computed on the input text, disregarding the special tokens or instruction text in the prompt. \texttt{KernelSHAP} samples instances in the neighborhood of $x$ to approximate Shapley Values. In the same way as \texttt{DeepLift}, we only perturb input tokens belonging to the input text, disregarding the rest of the prompt. Therefore, attribution is only computed on tokens from the instance of interest. Appendix~\ref{sec:appendix_post_hoc} provides more details about post hoc attribution implementation. 

The $k$ tokens with the highest attribution score are then selected to build the rationale: it is defined following the template \textit{"The $k$ keywords $\langle word_{1} \rangle$, $\langle word_{2}\rangle$,..., and $\langle word_{k} \rangle$ are important to predict that the answer is $\langle y \rangle$"}. This way, \method\ generates rationales from post hoc attribution methods by converting a numerical vector of importance into a natural language rationale.

\texttt{Self\_topk} consists in directly prompting $f$ to generate the $k$ most important tokens used to make its prediction. \texttt{Self\_topk} is generated in a \textit{predict-then-explain} post hoc manner, since the text containing the $k$ most important keywords is generated given the ground truth answer~$y$. 

Finally, \texttt{Ph-CoT} consists in prompting $f$ to generate a \textit{p}-step free text explanation in a post hoc manner, given the ground truth $y$. Therefore, \texttt{Ph-CoT} can be defined as a post hoc Chain-of-Thought explanation. The final related rationale $r$ is defined following the template \textit{"p-step rationale: $\langle \phi \rangle$ , therefore the answer is $\langle y \rangle$"}, where $\phi$ is the post-hoc free text rationale previously generated, and $p$ is the number of steps in the rationale. Appendix~\ref{sec:appendix_prompt} provides more details about the prompts used to generate \texttt{Self\_topk} and \texttt{Ph-CoT} rationales. We give several examples of generated rationales and answers conditioned by different rationale generator in Appendix~\ref{sec:appendix_examples}.

\subsection{Prompt Design for SLMs}
The final step consists in designing the prompt that is used to make the prediction on the test set. 

We define a \texttt{preprompt} at the beginning of the final prompt to define the instruction asked to $f$, i.e. generating a rationale and an answer to a specific question.  The \texttt{preprompt} can take two different forms, depending on the format of the generated rationales (\texttt{top\_{k}} important words or \textit{p}-step natural language explanation). More details about the \texttt{preprompt} are provided in Appendix~\ref{sec:appendix_prompt}.

Finally, the output prompt is built based on the previously generated rationales. The latter is built following the template: \textit{"\texttt{preprompt}, $(x_{1}, r_{1}, y_{1})$, $(x_{2}, r_{2}, y_{2})$, ..., $(x_{n}, r_{n}, y_{n})$"}. Finally, this \textit{n}-shot prompt is used as a context to make predictions in an ICL setting on the test set. 

\section{Experimental Settings} 
\label{xp}

\begin{table*}[t]
\centering
\small	
\begin{tabular}{cccccccc}
\hline
\multicolumn{1}{|c|}{\textbf{\begin{tabular}[c]{@{}c@{}}Model\\ (size)\end{tabular}}}                             & \multicolumn{1}{c|}{\textbf{Dataset}}                                                            & \multicolumn{1}{c|}{\textbf{\begin{tabular}[c]{@{}c@{}}Selection\\ strategy\end{tabular}}} & \textbf{\texttt{IO (ref.)}} & \textbf{\texttt{Auto-CoT}} & \multicolumn{1}{c|}{\textbf{\texttt{AMPLIFY}}}    & \multicolumn{2}{c|}{\textbf{\texttt{Self-AMPLIFY} (ours)}}                 \\
\multicolumn{1}{|l|}{}                                                                                            & \multicolumn{1}{c|}{}                                                                            & \multicolumn{1}{c|}{}                                                                      &                             &                            & \multicolumn{1}{c|}{\textbf{\texttt{BERT} proxy}} & \textbf{\texttt{DeepLift}} & \multicolumn{1}{c|}{\textbf{\texttt{Ph-CoT}}} \\ \hline
\multicolumn{1}{|c|}{\multirow{10}{*}{\textbf{\begin{tabular}[c]{@{}c@{}}\texttt{Mistral} \\ (7B)\end{tabular}}}} & \multicolumn{1}{c|}{\multirow{2}{*}{\begin{tabular}[c]{@{}c@{}}ARC\\ Challenge\end{tabular}}}    & \multicolumn{1}{c|}{\texttt{Success}}                                                      & 72.8                        & 71.8                       & \multicolumn{1}{c|}{70.4}                         & 71.1                       & \multicolumn{1}{c|}{\textbf{75.2*}}           \\
\multicolumn{1}{|c|}{}                                                                                            & \multicolumn{1}{c|}{}                                                                            & \multicolumn{1}{c|}{\texttt{Error}}                                                        & 69.0                        & 69.3                       & \multicolumn{1}{c|}{70.4}                         & 70.0                       & \multicolumn{1}{c|}{\textbf{72.8*}}           \\ \cline{2-8} 
\multicolumn{1}{|c|}{}                                                                                            & \multicolumn{1}{c|}{\multirow{2}{*}{\begin{tabular}[c]{@{}c@{}}Causal \\ Judgment\end{tabular}}} & \multicolumn{1}{c|}{\texttt{Success}}                                                      & 36.8                        & \textbf{63.2***}           & \multicolumn{1}{c|}{52.6***}                      & 52.6***                    & \multicolumn{1}{c|}{50.0***}                  \\
\multicolumn{1}{|c|}{}                                                                                            & \multicolumn{1}{c|}{}                                                                            & \multicolumn{1}{c|}{\texttt{Error}}                                                        & 31.6                        & 50.0**                     & \multicolumn{1}{c|}{39.5}                         & 55.3***                    & \multicolumn{1}{c|}{\textbf{60.5***}}         \\ \cline{2-8} 
\multicolumn{1}{|c|}{}                                                                                            & \multicolumn{1}{c|}{\multirow{2}{*}{CQA}}                                                        & \multicolumn{1}{c|}{\texttt{Success}}                                                      & 60.7                        & 61.3                       & \multicolumn{1}{c|}{60.7}                         & 66.7**                     & \multicolumn{1}{c|}{\textbf{67.6***}}         \\
\multicolumn{1}{|c|}{}                                                                                            & \multicolumn{1}{c|}{}                                                                            & \multicolumn{1}{c|}{\texttt{Error}}                                                        & 61.7                        & 59.3                       & \multicolumn{1}{c|}{64.7}                         & 62.3                       & \multicolumn{1}{c|}{\textbf{66.3*}}           \\ \cline{2-8} 
\multicolumn{1}{|c|}{}                                                                                            & \multicolumn{1}{c|}{\multirow{2}{*}{SIQA}}                                                       & \multicolumn{1}{c|}{\texttt{Success}}                                                      & 57.3                        & 60.0                       & \multicolumn{1}{c|}{56.0}                         & 59.7                       & \multicolumn{1}{c|}{\textbf{62.7**}}          \\
\multicolumn{1}{|c|}{}                                                                                            & \multicolumn{1}{c|}{}                                                                            & \multicolumn{1}{c|}{\texttt{Error}}                                                        & 59.3                        & 55.3                       & \multicolumn{1}{c|}{62.7*}                        & 61.7                       & \multicolumn{1}{c|}{\textbf{63.0*}}           \\ \cline{2-8} 
\multicolumn{1}{|c|}{}                                                                                            & \multicolumn{1}{c|}{\multirow{2}{*}{Snarks}}                                                     & \multicolumn{1}{c|}{\texttt{Success}}                                                      & 50.0                        & \textbf{66.7*}             & \multicolumn{1}{c|}{55.6}                         & 58.3                       & \multicolumn{1}{c|}{63.9*}                    \\
\multicolumn{1}{|c|}{}                                                                                            & \multicolumn{1}{c|}{}                                                                            & \multicolumn{1}{c|}{\texttt{Error}}                                                        & 36.1                        & 50.0*                      & \multicolumn{1}{c|}{47.2}                         & 52.8**                     & \multicolumn{1}{c|}{\textbf{72.2***}}         \\ \hline
\multicolumn{1}{l}{}                                                                                              & \multicolumn{1}{l}{}                                                                             & \multicolumn{1}{l}{}                                                                       & \multicolumn{1}{l}{}        & \multicolumn{1}{l}{}       & \multicolumn{1}{l}{}                              & \multicolumn{1}{l}{}       & \multicolumn{1}{l}{}                          \\ \hline
\multicolumn{1}{|c|}{\multirow{10}{*}{\textbf{\begin{tabular}[c]{@{}c@{}}\texttt{Zephyr}\\ (7B)\end{tabular}}}}   & \multicolumn{1}{c|}{\multirow{2}{*}{\begin{tabular}[c]{@{}c@{}}ARC\\ Challenge\end{tabular}}}    & \multicolumn{1}{c|}{\texttt{Success}}                                                      & 63.6                        & 63.3                       & \multicolumn{1}{c|}{67.0*}                        & 66.0                       & \multicolumn{1}{c|}{\textbf{70.7***}}         \\
\multicolumn{1}{|c|}{}                                                                                            & \multicolumn{1}{c|}{}                                                                            & \multicolumn{1}{c|}{\texttt{Error}}                                                        & 65.3                        & 65.6                       & \multicolumn{1}{c|}{\textbf{71.1**}}              & 68.4*                      & \multicolumn{1}{c|}{68.0}                     \\ \cline{2-8} 
\multicolumn{1}{|c|}{}                                                                                            & \multicolumn{1}{c|}{\multirow{2}{*}{\begin{tabular}[c]{@{}c@{}}Causal \\ Judgment\end{tabular}}} & \multicolumn{1}{c|}{\texttt{Success}}                                                      & 39.5                        & 55.3**                     & \multicolumn{1}{c|}{50.0}                         & 52.6*                      & \multicolumn{1}{c|}{\textbf{57.9**}}          \\
\multicolumn{1}{|c|}{}                                                                                            & \multicolumn{1}{c|}{}                                                                            & \multicolumn{1}{c|}{\texttt{Error}}                                                        & 42.1                        & 50.0                       & \multicolumn{1}{c|}{\textbf{60.5**}}              & 47.3                       & \multicolumn{1}{c|}{52.6*}                    \\ \cline{2-8} 
\multicolumn{1}{|c|}{}                                                                                            & \multicolumn{1}{c|}{\multirow{2}{*}{CQA}}                                                        & \multicolumn{1}{c|}{\texttt{Success}}                                                      & 53.3                        & 61.0***                    & \multicolumn{1}{c|}{61.3***}                      & \textbf{64.7***}           & \multicolumn{1}{c|}{62.3***}                  \\
\multicolumn{1}{|c|}{}                                                                                            & \multicolumn{1}{c|}{}                                                                            & \multicolumn{1}{c|}{\texttt{Error}}                                                        & 56.3                        & 63.0**                     & \multicolumn{1}{c|}{\textbf{68.0***}}             & 63.3**                     & \multicolumn{1}{c|}{66.7***}                  \\ \cline{2-8} 
\multicolumn{1}{|c|}{}                                                                                            & \multicolumn{1}{c|}{\multirow{2}{*}{SIQA}}                                                       & \multicolumn{1}{c|}{\texttt{Success}}                                                      & 53.7                        & 59.7**                     & \multicolumn{1}{c|}{56.7}                         & 59.0**                     & \multicolumn{1}{c|}{\textbf{65.0***}}         \\
\multicolumn{1}{|c|}{}                                                                                            & \multicolumn{1}{c|}{}                                                                            & \multicolumn{1}{c|}{\texttt{Error}}                                                        & 51.0                        & 60.0***                    & \multicolumn{1}{c|}{59.3***}                      & \textbf{60.3***}           & \multicolumn{1}{c|}{54.3*}                    \\ \cline{2-8} 
\multicolumn{1}{|c|}{}                                                                                            & \multicolumn{1}{c|}{\multirow{2}{*}{Snarks}}                                                     & \multicolumn{1}{c|}{\texttt{Success}}                                                      & 36.1                        & \textbf{44.4*}             & \multicolumn{1}{c|}{\textbf{44.4*}}               & 41.7                       & \multicolumn{1}{c|}{38.9}                     \\
\multicolumn{1}{|c|}{}                                                                                            & \multicolumn{1}{c|}{}                                                                            & \multicolumn{1}{c|}{\texttt{Error}}                                                        & 47.2                        & 41.7                       & \multicolumn{1}{c|}{41.7}                         & 52.8                       & \multicolumn{1}{c|}{\textbf{55.6*}}           \\ \hline
\end{tabular}
\vspace*{0.5em} 
\caption{\label{tab:results} 
\method\ and competitors accuracy (\%) on five test sets and two 7 billion parameters models. \method\ is tested on 2 versions, depending on the post hoc explainer used to generate rationales. \texttt{IO} stands for "input-output" standard prompting. \texttt{Auto-CoT} and \texttt{AMPLIFY} are two competing methods automatically generating rationales to enhance the input prompt. The best results are highlighted in bold. With $p$ as the $p$-value of the one-tailed paired $t$-test, *$p<10$\%, **$p<5$\%, ***$p<1$\%. \texttt{IO (ref.)} stands for the reference baseline.}
\end{table*}

This section presents the experimental study conducted across five datasets and three autoregressive SLMs of various size. We start by running two versions of \method\ on two 7 billion parameters, respectively based on two post hoc explainers (\texttt{DeepLift} and \texttt{Ph-CoT}). We compare these two versions of \method\ to \texttt{Auto-CoT} and \texttt{AMPLIFY}, two competitors automatically generating rationales and \texttt{IO} (Input-Output), a traditional prompting setup baseline. Next, we deeply assess the impact of the $topk$ post hoc explainers on \method\ performance through an ablation study. Finally, we run \method\ and the \texttt{Gemma} SLM in its 2 and 7 billion parameter versions and highlight the limits of our approach on tiny models.

\subsection{Experimental protocol.}
\paragraph{Datasets.}
\method\ is tested on five common LMs' benchmarks. ARC Challenge~\cite{arc}, CommonsenseQA (CQA)~\cite{CQA} and Social IQa (SIQA)~\cite{SIQA} are commonsense reasoning datasets requiring the ability to use prior knowledge about the world. The Snarks and Causal Judgment datasets~\cite{beyond_imitation_game} are datasets related to challenging complex tasks. Snarks requires to distinguish between sarcastic and non-sarcastic sentences, and Causal Judgment is designed to assess the ability in deducing causal factors from a detailed summary. These datasets are commonly used to evaluate LMs' performance. 

\paragraph{Models.}
We test \method\ on Instruction-tuned SLMs whose size does not exceed 7 billion parameters and achieve good results in common benchmarks. \texttt{Mistral-7B}~\cite{jiang2023mistral}, \texttt{Zephyr-7B}~\cite{tunstall_zephyr_2023} and \texttt{Gemma-7B}~\cite{gemma} are 7 billion parameters SLMs achieving state-of-the-art performance among other SLMs in a wide variety of NLP tasks. We then test the limits of \texttt{Self-AMPLIFY} on the smaller 2 billion parameter SLM \texttt{Gemma-2b} achieving strong performance for its size but with less reasoning abilities. 

\paragraph{\method\ versions and competitors.}

\begin{table*}[]
\small
\centering
\begin{tabular}{|cc|cccc|}
\hline
\multicolumn{1}{|c|}{\textbf{Dataset}}                                                                     & \textbf{\begin{tabular}[c]{@{}c@{}}Selection\\ strategy\end{tabular}} & \multicolumn{4}{c|}{\textbf{\texttt{Self-AMPLIFY}}}                                                                                      \\
\multicolumn{1}{|c|}{}                                                                                     &                                                                       & \textbf{\texttt{KernelShap}} & \textbf{\texttt{DeepLift}} & \multicolumn{1}{c|}{\textbf{\texttt{Self\_topk}}} & \textbf{\texttt{random}} \\ \hline
\multicolumn{1}{|c|}{\multirow{2}{*}{\textbf{\begin{tabular}[c]{@{}c@{}}ARC\\ Challenge\end{tabular}}}}    & \textbf{\texttt{Success}}                                             & \textbf{69.0 $\pm$ 2.3}      & 67.7 $\pm$ 2.0             & \multicolumn{1}{c|}{67.9 $\pm$ 2.7}               & 64.9 $\pm$ 5.0           \\
\multicolumn{1}{|c|}{}                                                                                     & \textbf{\texttt{Error}}                                               & 67.6 $\pm$ 2.7               & 67.7 $\pm$ 2.9             & \multicolumn{1}{c|}{\textbf{69.3 $\pm$ 1.6}}      & 63.7 $\pm$ 4.4           \\ \hline
\multicolumn{1}{|c|}{\multirow{2}{*}{\textbf{\begin{tabular}[c]{@{}c@{}}Causal \\ Judgment\end{tabular}}}} & \textbf{\texttt{Success}}                                             & 49.2 $\pm$ 5.0               & \textbf{49.7 $\pm$ 4.9}    & \multicolumn{1}{c|}{48.2 $\pm$ 5.6}               & 45.8 $\pm$ 4.1           \\
\multicolumn{1}{|c|}{}                                                                                     & \textbf{\texttt{Error}}                                               & 47.4 $\pm$ 5.1               & \textbf{52.6 $\pm$ 5.0}    & \multicolumn{1}{c|}{\textbf{52.6 $\pm$ 6.1}}      & 47.3 $\pm$ 6.0           \\ \hline
\multicolumn{1}{|c|}{\multirow{2}{*}{\textbf{CQA}}}                                                        & \textbf{\texttt{Success}}                                             & 58.8 $\pm$ 2.1               & 58.9 $\pm$ 1.8             & \multicolumn{1}{c|}{\textbf{59.9 $\pm$ 2.8}}      & 58.4 $\pm$ 2.6           \\
\multicolumn{1}{|c|}{}                                                                                     & \textbf{\texttt{Error}}                                               & 58.4 $\pm$ 2.9               & 59.7 $\pm$ 3.1             & \multicolumn{1}{c|}{\textbf{61.1 $\pm$ 2.8}}      & 57.2 $\pm$ 4.1           \\ \hline
\multicolumn{1}{|c|}{\multirow{2}{*}{\textbf{Snarks}}}                                                     & \textbf{\texttt{Success}}                                             & \textbf{50.6 $\pm$ 7.9}      & 50.0 $\pm$ 4.7             & \multicolumn{1}{c|}{49.7 $\pm$ 6.1}               & 50.0 $\pm$ 6.1           \\
\multicolumn{1}{|c|}{}                                                                                     & \textbf{\texttt{Error}}                                               & \textbf{53.6 $\pm$ 8.0}      & 52.0 $\pm$ 5.1             & \multicolumn{1}{c|}{50.0 $\pm$ 5.4}               & 47.8 $\pm$ 5.4           \\ \hline
\multicolumn{1}{|c|}{\multirow{2}{*}{\textbf{SIQA}}}                                                       & \textbf{\texttt{Success}}                                             & \textbf{59.6 $\pm$ 3.0}      & 59.0 $\pm$ 4.1             & \multicolumn{1}{c|}{58.6 $\pm$ 2.5}               & 54.9 $\pm$ 2.9           \\
\multicolumn{1}{|c|}{}                                                                                     & \textbf{\texttt{Error}}                                               & \textbf{60.6 $\pm$ 3.1}      & 60.0 $\pm$ 2.7             & \multicolumn{1}{c|}{58.7 $\pm$ 4.0}               & 55.9 $\pm$ 2.8           \\ \hline
\multicolumn{2}{|c|}{\textit{Average}}                                                                                                                                             & \textit{57,5}                & \textit{57,7}              & \multicolumn{1}{c|}{\textit{57,6}}                & \textit{54,6}            \\ \hline
\end{tabular}
\caption{\label{tab:attr_ablation}
Average accuracy (\%) and standard deviation computed on 10 \method\ runs for different $topk$ post hoc explainers on \texttt{Mistral-7B}.}
\end{table*}

We test four instantiations of the \method\ framework based on the four following  post hoc explanation methods: \texttt{DeepLift}, \texttt{KernelShap}, \texttt{Self\_topk} and \texttt{Ph-CoT}. In particular, we compared the two \texttt{DeepLift} and \texttt{Ph-CoT} instanciations of \method\ to a traditional ($x$, $y$) prompting setup (input-output, {\texttt{IO}), \texttt{Auto-CoT}~\cite{zhang_automatic_2022} and \texttt{AMPLIFY}~\cite{krishna_post_2023}. For a fair comparison, we run \method, \texttt{Auto-CoT} and \texttt{AMPLIFY} with the same $n$-shot sample context. This way, we focus our comparative analysis on the ability of each method to generate high quality rationales leading to accuracy improvement. \texttt{AMPLIFY} rationales are generated by applying \texttt{DeepLift} to a fine tuned \texttt{BERT-base} model. We give more details about the \texttt{AMPLIFY} implementation in Appendix~\ref{sec:appendix_slm_implementation_details}.


\method\ and its competitors are tested on the same corpora. Therefore, contexts are enriched from the same training corpus $\mathcal{T}$ and inference is performed on the same test sets. Finally, the output is retrieved in the correct format by using the assessed SLM as in~\citet{zhang_automatic_2022} to fit the label space (for example A or B for Snarks) and to compute accuracy. Because of the high computational cost of testing, we vary the number of runs and the size of test sets according to the performed analysis. We detail sample sizes, number of runs, context size ($n$), number of keywords ($k$) and number of steps ($p$) associated with each SLM and dataset in our experiments in Appendix~\ref{sec:appendix_exp_prot}. We present an in-depth analysis of the impact of $n$ and $k$ on \method\ performance in Appendix~\ref{sec:appendix_topk_context_size}.

\subsection{Results.}
\label{result}

\paragraph{Global Results.}
Table~\ref{tab:results} shows the experimental results obtained on \texttt{Mistral-7B} and \texttt{Zephyr-7B} by running once \method\ with \texttt{DeepLift} and \texttt{Ph-CoT} and its competitors on the same train set for rationale generation and the same test set for performance assessment. For each dataset/model case, one  of the \texttt{Self-AMPLIFY} modalities leads to the best result (for example \texttt{Ph-CoT} related to the \texttt{success} strategy for the \texttt{Mistral}/ARC Challenge case). The two \texttt{Self-AMPLIFY} modalities perform almost always better than the classical \texttt{IO} prompting and lead in average to higher accuracy as compared to \texttt{Auto-CoT}. The two instanciations of \method\ perform better than \texttt{AMPLIFY} with \texttt{Mistral-7B} without the use of a proxy additional fine-tuned model to generate rationales. In particular, post hoc \texttt{Ph-CoT} rationales give in average the best \method\ results. These results confirm the interest of \texttt{Self-AMPLIFY} to improve SLMs' performance fully automatically.

\paragraph{Impact of the selection strategy.}
Table~\ref{tab:results} highlights that the \texttt{success} selection strategy of \texttt{Self-AMPLIFY} gives good results overall, doing on average as well as the \texttt{error} one. This result confirms the interest of adding initially correctly classified examples into the context, which is not possible in the initial \texttt{AMPLIFY} framework. \texttt{Self-AMPLIFY} gives almost always better results than \texttt{AMPLIFY} when applied with the \texttt{success} strategy. However, \texttt{AMPLIFY} and \method\ have similar overall results with the \texttt{error} strategy. The results do not show whether a given selection strategy gives better results with \method.

\begin{figure*}[h]{\centering}
\begin{center}
\includegraphics[scale=0.32]{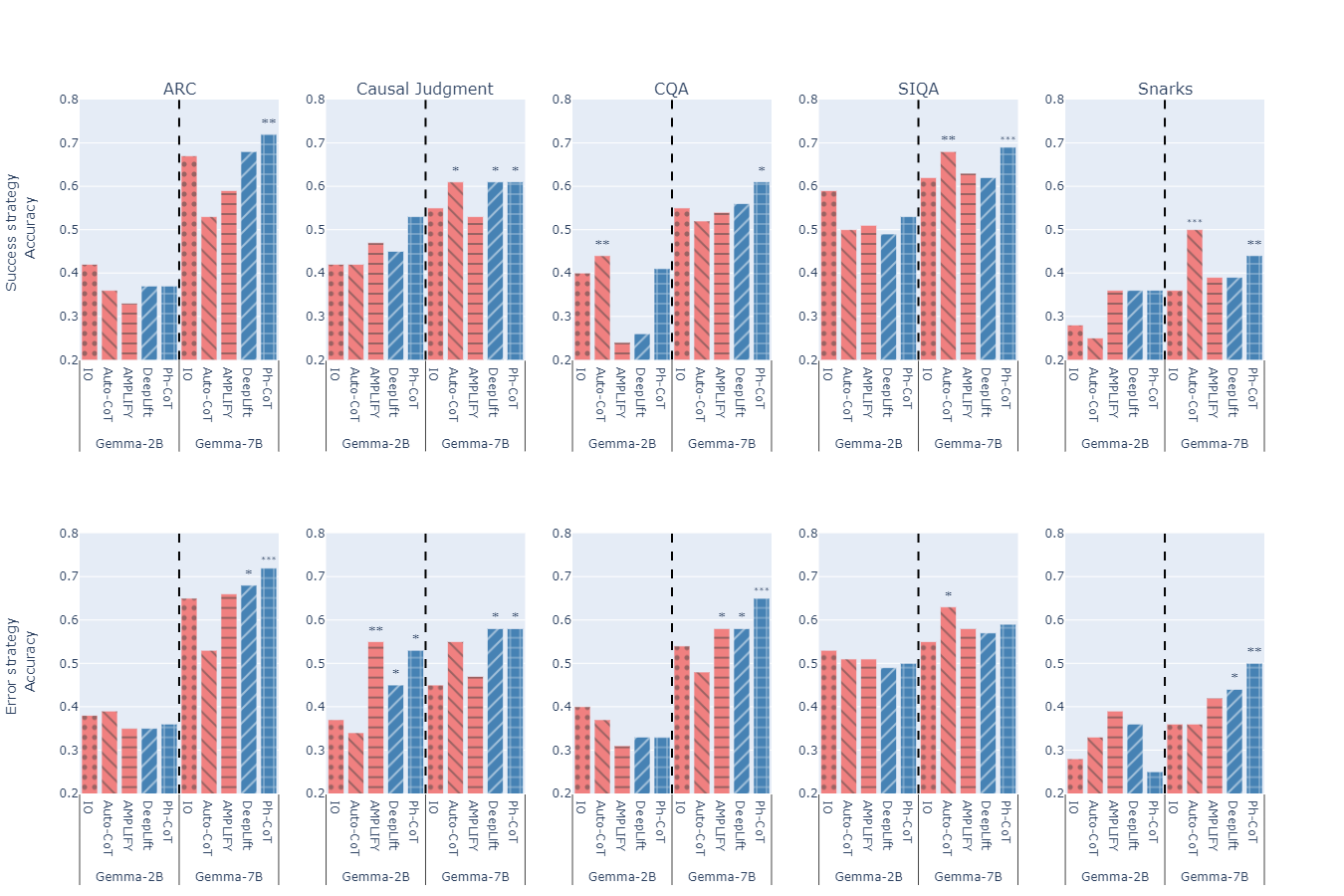}
\caption{\method\ (blue) and competitors (red) accuracy (\%) with Gemma-2 (left) and Gemma-7B (right). \method\ is run on 2 versions: \texttt{DeepLift} and \texttt{Ph-CoT}. With $p$ as the $p$-value of the paired $t$-test, *$p<10$\%, **$p<5$\%, ***$p<1$\%. \texttt{IO} stands for the reference baseline.}
\label{fig:gemma_2_7_B}
\end{center}
\end{figure*}

\paragraph{Impact of the post hoc explainer.}
Table~\ref{tab:results} shows that \texttt{Ph-CoT} post hoc explanations give in average the best \texttt{Self-AMPLIFY} results as compared to \texttt{DeepLift}. In particular, \texttt{Ph-CoT} related to the \texttt{error} strategy leads to significant highest performance gain compared to other competitors for complex tasks such as Snarks and Causal Judgment. We hypothesize that this is linked to the ability of SLMs to generate faithful free text natural language post hoc explanations as a corrective signal.  Table~\ref{tab:attr_ablation} shows the results of the ablation study of the $topk$ explanations on \texttt{Mistral-7B} obtained on 10 \method\ runs. The \texttt{random} $topk$ rationale generator gives the worst accuracy compare to the other \texttt{Self-AMPLIFY} instantiations. This result shows that the rationale signal content can have an impact on the \method\ performance. The different $topk$ instantiations of \method\ give very similar results on average, indicating that the framework is robust.

Based on these results, the default setting of \method\ $topk$ explainer only depends on whether or not the model parameters are accessible. If \texttt{DeepLift} is much less computationally costly than \texttt{KernelShap}, it can only be applied if the model's internal parameters are accessible, which is not always the case with online APIs. \texttt{Self\_topk} is less costly than \texttt{DeepLift} in that it is only based on text generation, without any additional computation. However, it is difficult to completely control the format of the $topk$ explanations, as text generation does not always respect the template initially provided.

\paragraph{The size limit of post hoc rationale enhancement.} 

We extend our analysis with two new SLMs: Gemma-7B and the tiny Gemma-2B. Figure~\ref{fig:gemma_2_7_B} shows that on average, \method\ outperforms its competitors on \texttt{Gemma-7B} in the same way as with \texttt{Zephyr-7B} and \texttt{Mistral-7B}. Every version of \method\ consistently outperforms \texttt{IO} and \texttt{AMPLIFY} and do better on average than \texttt{Auto-CoT}. However, rationale enhancement leads to much poorer results with \texttt{Gemma-2B} as compared to \texttt{Gemma-7B}: \method, \texttt{Auto-CoT} and \texttt{AMPLIFY} do barely as well on average than the \texttt{IO} baseline. We hypothesize that these contrasting results are linked to \texttt{Gemma-2B}'s small size and less advanced reasoning capabilities as compared to 7-billion parameters models. \texttt{Gemma}'s results are presented with those obtained with \texttt{Zephyr-7B} and \texttt{Mistral-7B} in Appendix~\ref{sec:appendix_exp_prot} in Table~\ref{tab:results_all}.

\section{Discussion}
\label{sec:disc}

The \method\ framework is versatile and can work with any other post hoc attribution methods, such as Integrated Gradient or LIME. We recommend as \method\ default setting \texttt{Ph-CoT} rationales if the aim is only to obtain the most accurate results. However, a framework user might expect the generated rationales to be faithful to build trust with \method~\cite{trust_xai}. Free text rationale faithfulness evaluation is a difficult task and there is no consensus in the way to measure it~\cite{wiegreffe2021measuring, atanasova_faithfulness_2023}. Faithfulness assessment is easier with $topk$ rationales by computing common metrics such as stability~\cite{faithful_xai} and self-consistency~\cite{self_consistency_madsen}. Therefore, \texttt{KernelShap}, \texttt{DeepLift} or other post hoc attribution explainer should be preferred if rationale faithfulness evaluation is needed. The appropriate $topk$ explainer can then be chosen depending on the level of information available about the model as stated in the previous section.

As a future work, \method\ could be improved by generating other types of rationales to enrich the prompt such as counterfactual examples (see~\citet{tigtec} for a recent method). A deeper analysis of the link between task complexity, selection strategy and \method\ performance would also provide information on how to better generate valuable rationales. Finally, it would be enlightening to assess the faithfulness of \method\ generated rationales. For instance, ICL generated rationales could be compared to ground truth explanations obtained in a post hoc manner. We see these perspectives as promising paths towards a better understanding of LMs' ability to faithfully learn to self-explain. 

\section{Conclusion}
\label{conclusion}
We introduced \method, an extension of the \texttt{AMPLIFY} framework, automatically generating rationales to enrich the prompt in ICL settings for SLMs. \method\ is the first approach enriching the prompt without human-annotated rationales or the use of auxiliary models, but only with the SLM itself. \method\ implements 2 selection strategies and 4 post hoc explanation methods, making it versatile and flexible. \method\ results in performance gain compared to its competitors in the considered tasks for 7 billion parameters models. Finally, this work sheds light on the interest of using post hoc explanations to enhance SLM's performance. 

\section{Limitations}
\paragraph{Datasets and models.}
In this work we have tested \method\ by applying it on 5 datasets and 3 language models. The conclusions of our work would have more weight if other models were included in the study. Furthermore, it would be interesting to apply \method\ on slightly bigger SLMs with better reasoning abilities. This would make the framework even more useful to the community.

\paragraph{Rationale relevance.}
The quality of the generated rationales is not assessed, neither when enriching the prompt (rationale generation step), nor when generating the text (prediction on the test set). These rationales should be interpreted with caution, as they have been generated solely to enhance SLMs' performance. This phenomenon has been raised by~\cite{zhang_automatic_2022_old}, where wrong demonstrations based on low quality rationales can still lead to performance gains.

\paragraph{Computational cost.}
The use of KernelShap and DeepLift is computationally costly. Even if it is affordable to use them with SLMs, the resource requirement is substantial. One could lower the number of samples used to compute KernelShap if needed (see Appendix~\ref{sec:appendix_post_hoc}) to make it even more affordable.

\label{limitations}

\section*{Ethics Statement}
Since SLMs' training data can be biased, there is a risk of generating harmful text during inference. One using \method\ to generate rationales must be aware of these biases in order to stand back and analyze the produced texts. Finally, SLMs consume energy, potentially emitting greenhouse gases. They must be used with caution.  

\bibliography{amplify}
\bibliographystyle{acl_natbib}
\appendix
\section{Appendix}
\label{sec:appendix}
\subsection{Scientific libraries}
We used several open-source libraries in this work: pytorch~\cite{paszke2019pytorch}, HuggingFace transformers~\cite{wolf2020transformers} sklearn~\cite{pedregosa2011scikit} and Captum~\cite{miglani_using_2023}. 

\subsection{SLMs implementation Details}
\label{sec:appendix_slm_implementation_details}
\paragraph{Small Language Models.} The library used to import the pretrained SLMs is Hugging-Face. In particular, the backbone version of Mistral is \texttt{Mistral-7B-Instruct-v0.2} the one of Zephyr is \texttt{zephyr-7b-beta}, and the one of Gemma are respectively \texttt{gemma-1.1-7b-it} and \texttt{gemma-1.1-2b-it}. 

\paragraph{Instruction special tokens.}
The special tokens to use SLMs in instruction mode were the followings:
\begin{itemize}
    \item \texttt{Mistral-7B-Instruct-v0.2} \begin{itemize}
        \item \texttt{user\_token='[INST]'}
        \item \texttt{assistant\_token='[/INST]'}
        \item \texttt{stop\_token='</s>'}
    \end{itemize}
    \item \texttt{Zephyr-7b-beta} \begin{itemize}
        \item \texttt{user\_token='<|user|>'}
        \item \texttt{assistant\_token='<|assistant|>'}
        \item \texttt{stop\_token='</s>'}
    \end{itemize}
    \item \texttt{Gemma-1.1-2b-it} \begin{itemize}
        \item \texttt{user\_token= '<start\_of\_turn>user'}
        \item \texttt{assistant\_token= '<start\_of\_turn>model'}
        \item \texttt{stop\_token='<eos>'}
    \end{itemize}
    \item \texttt{Gemma-1.1-7b-it} \begin{itemize}
        \item \texttt{user\_token=' <start\_of\_turn>user'}
        \item \texttt{assistant\_token= '<start\_of\_turn>model'}
        \item \texttt{stop\_token='<eos>'}
    \end{itemize}
\end{itemize}
\paragraph{Text generation.}
Text generation was performed using the native functions of the Hugging Face library: \texttt{generate}. The \texttt{generate} function has been used with the following parameters:
\begin{itemize}
    \item \texttt{max\_new\_tokens = 300}
    \item \texttt{do\_sample = True}
    \item \texttt{num\_beams = 2}
    \item \texttt{no\_repeat\_ngram\_size = 2}
    \item \texttt{early\_stopping = True}
    \item \texttt{temperature = 0.95}
\end{itemize}

\paragraph{AMPLIFY implementation}
\texttt{AMPLIFY} has been implemented by fine tuning one \texttt{BERT-base} per training set. Table~\ref{tab::amplify_proxy} gives more information about \texttt{AMPLIFY} proxy models used to generate $topk$ explanations to enhance the prompt. The $topk$ post-hoc explanation method used was \texttt{DeepLift}. 

\begin{table}[h]
\footnotesize
\begin{tabular}{|c|c|c|c|}
\hline
\textbf{Dataset}                                          & \textbf{Accuracy (\%)} & \textbf{nb epoch} & \textbf{Proxy model} \\ \hline
\begin{tabular}[c]{@{}c@{}}ARC \\ Challenge\end{tabular}  & 25.7                  & 25                & \texttt{BERT-base}   \\ \hline
\begin{tabular}[c]{@{}c@{}}Causal\\ Judgment\end{tabular} & 52.6                  & 25                & \texttt{BERT-base}   \\ \hline
CQA                                                       & 20.9                  & 25                & \texttt{BERT-base}   \\ \hline
Snarks                                                    & 61.1                  & 25                & \texttt{BERT-base}   \\ \hline
SIQA                                                      & 56.8                  & 10                & \texttt{BERT-base}   \\ \hline
\end{tabular}
\caption{\label{tab::amplify_proxy} \texttt{AMPLIFY} proxy models performance and number of epochs by dataset.}
\end{table}

\subsection{Experimental protocol details}
\label{sec:appendix_exp_prot}

\begin{table}[h]
\scriptsize
\begin{tabular}{cc|ccc|}
\cline{3-5}
                                                                                                         &                                                                   & \multicolumn{3}{c|}{\textbf{Experiment}}                                                                                                                                                                                                                                                        \\ \cline{3-5} 
                                                                                                         &                                                                   & \multicolumn{1}{c|}{\textbf{\begin{tabular}[c]{@{}c@{}}Mistral-7B \\ and \\ Zephyr-7B\end{tabular}}} & \multicolumn{1}{c|}{\textbf{\begin{tabular}[c]{@{}c@{}}Mistral-7B\\ $topk$\\ ablation study\end{tabular}}} & \textbf{\begin{tabular}[c]{@{}c@{}}Gemma-7B \\ and\\ Gemma-2B\end{tabular}} \\ \hline
\multicolumn{2}{|c|}{\textbf{\begin{tabular}[c]{@{}c@{}}Corresponding\\ table/figure\end{tabular}}}                                                                          & \multicolumn{1}{c|}{Table 1}                                                                         & \multicolumn{1}{c|}{Table 2}                                                                               & Figure 4                                                                    \\ \hline
\multicolumn{2}{|c|}{\textbf{Nb runs}}                                                                                                                                       & \multicolumn{1}{c|}{1}                                                                               & \multicolumn{1}{c|}{10}                                                                                    & 1                                                                           \\ \hline
\multicolumn{1}{|c|}{\multirow{5}{*}{\textbf{\begin{tabular}[c]{@{}c@{}}Test set \\ size\end{tabular}}}} & \textbf{\begin{tabular}[c]{@{}c@{}}ARC\\ Challenge\end{tabular}}  & \multicolumn{1}{c|}{295}                                                                             & \multicolumn{1}{c|}{150}                                                                                   & 295                                                                         \\ \cline{2-5} 
\multicolumn{1}{|c|}{}                                                                                   & \textbf{\begin{tabular}[c]{@{}c@{}}Causal\\ Jugment\end{tabular}} & \multicolumn{1}{c|}{36}                                                                              & \multicolumn{1}{c|}{36}                                                                                    & 36                                                                          \\ \cline{2-5} 
\multicolumn{1}{|c|}{}                                                                                   & \textbf{CQA}                                                      & \multicolumn{1}{c|}{300}                                                                             & \multicolumn{1}{c|}{150}                                                                                   & 300                                                                         \\ \cline{2-5} 
\multicolumn{1}{|c|}{}                                                                                   & \textbf{Snarks}                                                   & \multicolumn{1}{c|}{39}                                                                              & \multicolumn{1}{c|}{39}                                                                                    & 39                                                                          \\ \cline{2-5} 
\multicolumn{1}{|c|}{}                                                                                   & \textbf{SIQA}                                                     & \multicolumn{1}{c|}{300}                                                                             & \multicolumn{1}{c|}{150}                                                                                   & 300                                                                         \\ \hline
\end{tabular}
\caption{\label{tab:xp_protocol} Experimental protocols details. Number of runs and test set sizes vary depending on the performed analysis.}
\end{table}

\begin{table}[h]
\scriptsize
\begin{tabular}{|c|c|c|}
\hline
\textbf{Hyperparameter}                                                               & \textbf{\begin{tabular}[c]{@{}c@{}}Mistral-7B, Zephyr-7B, \\ Gemma-7B\end{tabular}} & \textbf{Gemma-2B} \\ \hline
\begin{tabular}[c]{@{}c@{}}Context size\\ $n$\end{tabular}                            & \begin{tabular}[c]{@{}c@{}}6 (Causal Judgment)\\ 8 (other datasets)\end{tabular}    & 4                 \\ \hline
\begin{tabular}[c]{@{}c@{}}Number of keywords\\ in topk rationales\\ $k$\end{tabular} & 6                                                                                   & 4                 \\ \hline
\begin{tabular}[c]{@{}c@{}}Number of steps \\ in Ph-CoT\\ $p$\end{tabular}            & 3                                                                                   & 3                 \\ \hline
\end{tabular}
\caption{\label{tab:xp_protocol_2} \method\ hyperparameters per model per dataset.}
\end{table}

Table~\ref{tab:xp_protocol} shows the experimental protocol details of the performed analysis. Test set size is 39 for Snarks and 36 for Causal Judgment for every experiment. However, test sets are obtained by randomly sampling for ARC Challenge, CQA and SIQA with a varying size. Number of runs can also vary from one experiment to another. This is due to the high computational cost of running \method\ and its competitors with various selection strategy modalities on such a high number of datasets and texts. Since the ablation study only concerns 3 \method\ modalities and a \texttt{random} baseline, experiment contains 10 runs. Test size is however smaller for ARC Challenge, CQA and SIQA.

Table~\ref{tab:xp_protocol_2} presents the hyperparameters of \method\ and the context size of the experiments. Post hoc attribution methods and \texttt{Self\_topk} are computed with $k=6$ for \texttt{Mistral-7B}, \texttt{Zephyr-7B} and \texttt{Gemma-7B} and $k=4$ for \texttt{Gemma-2B}. \texttt{Ph-CoT} $p$-step rationales are generated with $p=3$. ICL context size is set at $n=8$ for \texttt{Zephyr-7B}, \texttt{Mistral-7B} and \texttt{Gemma-7B} for all the datasets, except for Causal Judgment, where $n=6$. The ICL size is set at $n=4$ for \texttt{Gemma-2B}, this smaller model being less able to handle long contexts.

The results of the experimental protocol are all presented in Table~\ref{tab:results_all}.

\subsection{Prompting format}
\label{sec:appendix_prompt}
Here we provide some details of different prompts used to give instructions to SLMs. 
\\

\textbf{Prompt for \texttt{Self\_topk} rationale generation}
\\
\textbf{user} \\
\textit{Choose the right answer with the $\langle$\texttt{topk}$\rangle$ most important keywords used to answer. Example: The answer is (A), the $\langle$\texttt{topk}$\rangle$ most important keywords to make the prediction are "word$_{1}$", ... and "word$_{k}$"}
\\

\textbf{Preprompt for \texttt{Ph-CoT} rationale generation}
\\
\textbf{user} \\
\textit{Choose the right answer and generate a concise $\langle$\texttt{n\_steps}$\rangle$-step explanation, with only one sentence per step. Example: The answer is (A), $\langle$\texttt{n\_steps}$\rangle$-step explanation: step$_{1}$, step$_{2}$,...,step$_{n}$.}
\\

\textbf{Final ICL n-samples prompt example based on $topk$ rationales}
\\
\textbf{user}\\
\textit{You are presented with multiple choice question, where choices will look like (A), (B), (C) or (D), generate $\langle$\texttt{topk\_words}$\rangle$ keywords providing hints and generate the right single answer
Ouput example: The $\langle$\texttt{topk\_words}$\rangle$ keywords "word$_{1}$", "word$_{2}$" ... and "word$_{k}$"  are important to predict that the answer is (A)} \\
$\langle$ \texttt{question$_{1}$} $\rangle$ \\
\textbf{assistant} \\
$\langle$ \texttt{rationale$_{1}$} $\rangle$ \\
$\langle$ \texttt{answer$_{1}$} $\rangle$ \\
... \\
\textbf{user} \\
$\langle$ \texttt{question$_{n}$} $\rangle$ \\
\textbf{assistant} \\
$\langle$ \texttt{rationale$_{n}$} $\rangle$ \\
$\langle$ \texttt{answer$_{n}$} $\rangle$ \\
\textbf{user}\\
$\langle$ \texttt{question$_{n+1}$}$\rangle$


\subsection{Post hoc attribution explanation methods}
\label{sec:appendix_post_hoc}
\paragraph{Captum library.}
Post hoc attribution has been computed using the Captum~\cite{miglani_using_2023} library. \method\ implements additional post hoc attribution methods as compared to those presented in our paper. These additional post hoc attribution methods can be used in the \method\ framework to generate rationales. Overall, we implement the following methods: 
\begin{itemize}
    \item Gradient-based \begin{itemize}
        \item \texttt{GradientXActivation}
        \item \texttt{IntegratedGradients}
        \item \texttt{DeepLift}
        \end{itemize}
        \item Perturbation-based \begin{itemize}
            \item \texttt{FeatureAblation}
            \item \texttt{Lime}
            \item \texttt{KernelShap}
            \item \texttt{ShapleyValueSampling}
            \item \texttt{ShapleyValues}
        \end{itemize}
\end{itemize}

\paragraph{Attribution implementation details.}
In particular, gradient-based approach are computed with respect to the SLM embedding layer (\texttt{layer = model.model.embed\_tokens}).

The parameters used to computed \texttt{DeepLift} and \texttt{KernelShap} were Captum's default settings. In particular, \texttt{KernelShap} was computed with \texttt{n\_samples = 350}

\paragraph{Baseline choice.}
The baseline choice is decisive for DeepLift computation. The baseline is selected so that importance is only computed with respect to the initial prompt, so that special tokens and preprompt have a null attribution. The baseline is thus constructed as a modified version of the text on which DeepLift is applied. Therefore, the part of the baseline where the attribution must have a non-zero value (here statement, question and possible answer) is replaced with padding.

\subsection{Impact of $topk$ explanation length and context size}
\label{sec:appendix_topk_context_size}
Figure~\ref{fig:topk_ablation} shows the evolution of the accuracy of \method\ with respect to the $topk$ hyperparameter. It turns out that the $topk$ explanation length does not seem to have an impact on the accuracy. Every $topk$ value gives better results than \texttt{IO prompting}. Figure~\ref{fig:n_sensib} shows the evolution of the accuracy of \method\ and \texttt{IO} prompting with respect to context size. Evaluation is made with \texttt{Mistral} and \texttt{Zephyr} and the Causal Judgment dataset. Most context sizes result in better \method\ result as compared to \texttt{IO}.

\begin{figure*}[h]{\centering}
\begin{center}
\includegraphics[scale=0.40]{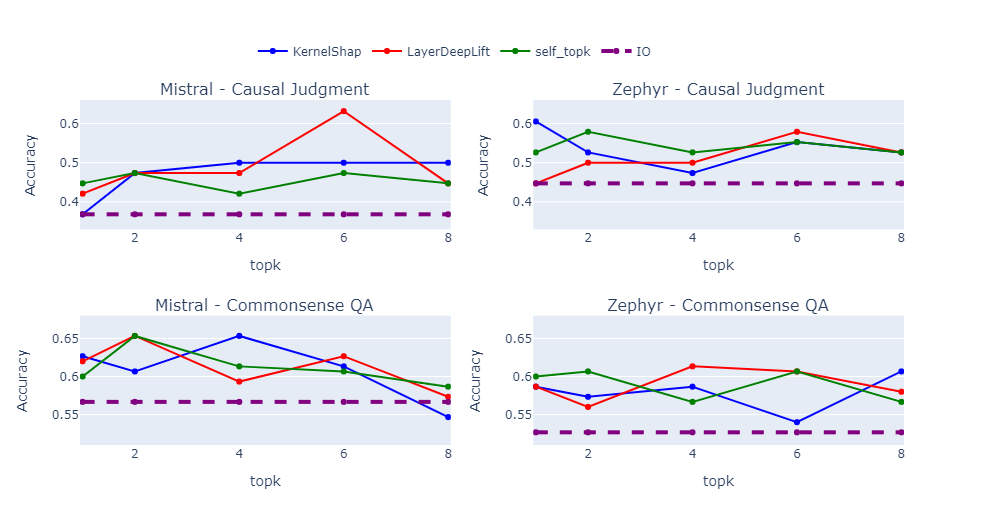}
\caption{\label{fig:topk_ablation}
Accuracy (\%) of classical IO prompting and \method\ for different $topk$ post hoc explainers and different \texttt{topk} values. Evaluation is made with \texttt{Mistral} and \texttt{Zephyr} on Commonsense QA and  Causal Judgment datasets.}

\end{center}
\end{figure*}

\begin{figure*}[h]{\centering}
\begin{center}
\includegraphics[scale=0.40]{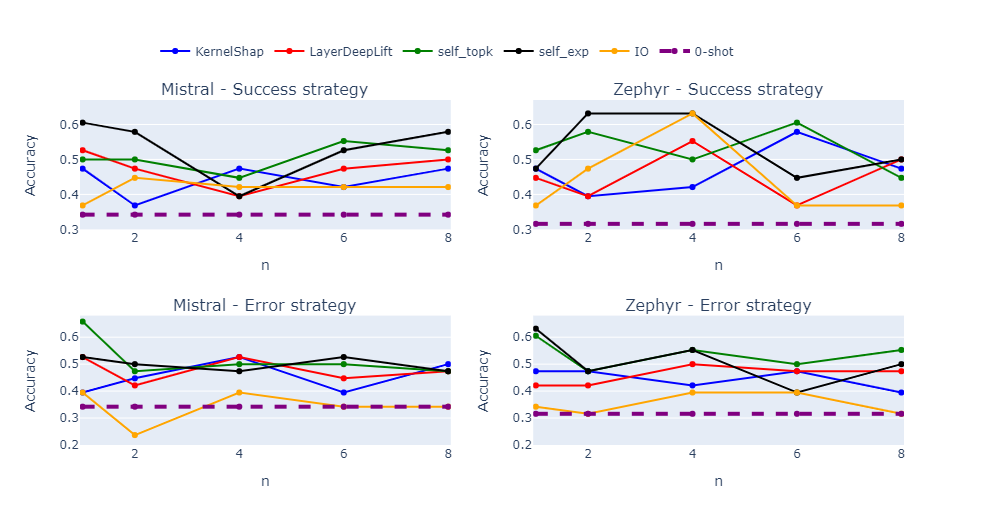}
\caption{\label{fig:n_sensib}
Accuracy (\%) of \method\ and classical IO prompting for different context sizes. Evaluation is made with \texttt{Mistral} and \texttt{Zephyr} for every selection strategy on the Causal Judgment dataset.}
\end{center}
\end{figure*}

\subsection{\method\ and competitors generated text example}
\label{sec:appendix_examples}
Figure~\ref{fig:text_example_arc},~\ref{fig:text_example_causal_judg},~\ref{fig:text_example_cqa},~\ref{fig:text_example_snarks} and \ref{fig:text_example_siqa} show several examples of generated texts conditioned by different rationale generators for every analyzed datasets.

\begin{table*}[t]
\centering
\small	
\begin{tabular}{cccccccc}
\hline
\multicolumn{1}{|c|}{\textbf{\begin{tabular}[c]{@{}c@{}}Model\\ (size)\end{tabular}}}                             & \multicolumn{1}{c|}{\textbf{Dataset}}                                                            & \multicolumn{1}{c|}{\textbf{\begin{tabular}[c]{@{}c@{}}Selection\\ strategy\end{tabular}}} & \textbf{\texttt{IO (ref.)}} & \textbf{\texttt{Auto-CoT}} & \multicolumn{1}{c|}{\textbf{\texttt{AMPLIFY}}}    & \multicolumn{2}{c|}{\textbf{\texttt{Self-AMPLIFY} (ours)}}                 \\
\multicolumn{1}{|l|}{}                                                                                            & \multicolumn{1}{c|}{}                                                                            & \multicolumn{1}{c|}{}                                                                      &                      &                            & \multicolumn{1}{c|}{\textbf{\texttt{BERT} proxy}} & \textbf{\texttt{DeepLift}} & \multicolumn{1}{c|}{\textbf{\texttt{Ph-CoT}}} \\ \hline
\multicolumn{1}{|c|}{\multirow{10}{*}{\textbf{\begin{tabular}[c]{@{}c@{}}\texttt{Mistral} \\ (7B)\end{tabular}}}} & \multicolumn{1}{c|}{\multirow{2}{*}{\begin{tabular}[c]{@{}c@{}}ARC\\ Challenge\end{tabular}}}    & \multicolumn{1}{c|}{\texttt{Success}}                                                      & 72.8                 & 71.8                       & \multicolumn{1}{c|}{70.4}                         & 71.1                       & \multicolumn{1}{c|}{\textbf{75.2*}}           \\
\multicolumn{1}{|c|}{}                                                                                            & \multicolumn{1}{c|}{}                                                                            & \multicolumn{1}{c|}{\texttt{Error}}                                                        & 69.0                 & 69.3                       & \multicolumn{1}{c|}{70.4}                         & 70.0                       & \multicolumn{1}{c|}{\textbf{72.8*}}           \\ \cline{2-8} 
\multicolumn{1}{|c|}{}                                                                                            & \multicolumn{1}{c|}{\multirow{2}{*}{\begin{tabular}[c]{@{}c@{}}Causal \\ Judgment\end{tabular}}} & \multicolumn{1}{c|}{\texttt{Success}}                                                      & 36.8                 & \textbf{63.2***}           & \multicolumn{1}{c|}{52.6***}                      & 52.6***                    & \multicolumn{1}{c|}{50.0***}                  \\
\multicolumn{1}{|c|}{}                                                                                            & \multicolumn{1}{c|}{}                                                                            & \multicolumn{1}{c|}{\texttt{Error}}                                                        & 31.6                 & 50.0**                     & \multicolumn{1}{c|}{39.5}                         & 55.3***                    & \multicolumn{1}{c|}{\textbf{60.5***}}         \\ \cline{2-8} 
\multicolumn{1}{|c|}{}                                                                                            & \multicolumn{1}{c|}{\multirow{2}{*}{CQA}}                                                        & \multicolumn{1}{c|}{\texttt{Success}}                                                      & 60.7                 & 61.3                       & \multicolumn{1}{c|}{60.7}                         & 66.7**                     & \multicolumn{1}{c|}{\textbf{67.6***}}         \\
\multicolumn{1}{|c|}{}                                                                                            & \multicolumn{1}{c|}{}                                                                            & \multicolumn{1}{c|}{\texttt{Error}}                                                        & 61.7                 & 59.3                       & \multicolumn{1}{c|}{64.7}                         & 62.3                       & \multicolumn{1}{c|}{\textbf{66.3*}}           \\ \cline{2-8} 
\multicolumn{1}{|c|}{}                                                                                            & \multicolumn{1}{c|}{\multirow{2}{*}{SIQA}}                                                       & \multicolumn{1}{c|}{\texttt{Success}}                                                      & 57.3                 & 60.0                       & \multicolumn{1}{c|}{{56.0}}                & 59.7                       & \multicolumn{1}{c|}{\textbf{62.7**}}          \\
\multicolumn{1}{|c|}{}                                                                                            & \multicolumn{1}{c|}{}                                                                            & \multicolumn{1}{c|}{\texttt{Error}}                                                        & 59.3                 & 55.3                       & \multicolumn{1}{c|}{62.7*}                        & 61.7                       & \multicolumn{1}{c|}{\textbf{63.0*}}           \\ \cline{2-8} 
\multicolumn{1}{|c|}{}                                                                                            & \multicolumn{1}{c|}{\multirow{2}{*}{Snarks}}                                                     & \multicolumn{1}{c|}{\texttt{Success}}                                                      & 50.0                 & \textbf{66.7*}             & \multicolumn{1}{c|}{55.6}                         & 58.3                       & \multicolumn{1}{c|}{63.9*}                    \\
\multicolumn{1}{|c|}{}                                                                                            & \multicolumn{1}{c|}{}                                                                            & \multicolumn{1}{c|}{\texttt{Error}}                                                        & 36.1                 & 50.0*                      & \multicolumn{1}{c|}{47.2}                         & 52.8**                     & \multicolumn{1}{c|}{\textbf{72.2***}}         \\ \hline
\multicolumn{1}{l}{}                                                                                              & \multicolumn{1}{l}{}                                                                             & \multicolumn{1}{l}{}                                                                       & \multicolumn{1}{l}{} & \multicolumn{1}{l}{}       & \multicolumn{1}{l}{}                              & \multicolumn{1}{l}{}       & \multicolumn{1}{l}{}                          \\ \hline
\multicolumn{1}{|c|}{\multirow{10}{*}{\textbf{\begin{tabular}[c]{@{}c@{}}\texttt{Zephyr}\\ (7B)\end{tabular}}}}   & \multicolumn{1}{c|}{\multirow{2}{*}{\begin{tabular}[c]{@{}c@{}}ARC\\ Challenge\end{tabular}}}    & \multicolumn{1}{c|}{\texttt{Success}}                                                      & 63.6                 & 63.3                       & \multicolumn{1}{c|}{67.0}                         & 66.0                       & \multicolumn{1}{c|}{\textbf{70.7***}}         \\
\multicolumn{1}{|c|}{}                                                                                            & \multicolumn{1}{c|}{}                                                                            & \multicolumn{1}{c|}{\texttt{Error}}                                                        & 65.3                 & 65.6                       & \multicolumn{1}{c|}{\textbf{71.1**}}              & 68.4*                      & \multicolumn{1}{c|}{68.0}                     \\ \cline{2-8} 
\multicolumn{1}{|c|}{}                                                                                            & \multicolumn{1}{c|}{\multirow{2}{*}{\begin{tabular}[c]{@{}c@{}}Causal \\ Judgment\end{tabular}}} & \multicolumn{1}{c|}{\texttt{Success}}                                                      & 39.5                 & 55.3**                     & \multicolumn{1}{c|}{50.0}                         & 52.6                       & \multicolumn{1}{c|}{\textbf{57.9**}}          \\
\multicolumn{1}{|c|}{}                                                                                            & \multicolumn{1}{c|}{}                                                                            & \multicolumn{1}{c|}{\texttt{Error}}                                                        & 42.1                 & 50.0                       & \multicolumn{1}{c|}{\textbf{60.5**}}              & 47.3                       & \multicolumn{1}{c|}{52.6*}                    \\ \cline{2-8} 
\multicolumn{1}{|c|}{}                                                                                            & \multicolumn{1}{c|}{\multirow{2}{*}{CQA}}                                                        & \multicolumn{1}{c|}{\texttt{Success}}                                                      & 53.3                 & 61.0***                    & \multicolumn{1}{c|}{61.3***}                      & \textbf{64.7***}           & \multicolumn{1}{c|}{62.3***}                  \\
\multicolumn{1}{|c|}{}                                                                                            & \multicolumn{1}{c|}{}                                                                            & \multicolumn{1}{c|}{\texttt{Error}}                                                        & 56.3                 & 63.0**                     & \multicolumn{1}{c|}{\textbf{68.0***}}             & 63.3**                     & \multicolumn{1}{c|}{66.7***}                  \\ \cline{2-8} 
\multicolumn{1}{|c|}{}                                                                                            & \multicolumn{1}{c|}{\multirow{2}{*}{SIQA}}                                                       & \multicolumn{1}{c|}{\texttt{Success}}                                                      & 53.7                 & 59.7**                     & \multicolumn{1}{c|}{56.7}                         & 59.0**                     & \multicolumn{1}{c|}{\textbf{65.0***}}         \\
\multicolumn{1}{|c|}{}                                                                                            & \multicolumn{1}{c|}{}                                                                            & \multicolumn{1}{c|}{\texttt{Error}}                                                        & 51.0                 & 60.0***                    & \multicolumn{1}{c|}{59.3***}                      & \textbf{60.3***}           & \multicolumn{1}{c|}{54.3*}                    \\ \cline{2-8} 
\multicolumn{1}{|c|}{}                                                                                            & \multicolumn{1}{c|}{\multirow{2}{*}{Snarks}}                                                     & \multicolumn{1}{c|}{\texttt{Success}}                                                      & 36.1                 & \textbf{44.4*}             & \multicolumn{1}{c|}{\textbf{44.4*}}               & 41.7                       & \multicolumn{1}{c|}{38.9}                     \\
\multicolumn{1}{|c|}{}                                                                                            & \multicolumn{1}{c|}{}                                                                            & \multicolumn{1}{c|}{\texttt{Error}}                                                        & 47.2                 & 41.7                       & \multicolumn{1}{c|}{41.7}                         & 52.8                       & \multicolumn{1}{c|}{\textbf{55.6*}}           \\ \hline
\multicolumn{1}{l}{}                                                                                              & \multicolumn{1}{l}{}                                                                             & \multicolumn{1}{l}{}                                                                       & \multicolumn{1}{l}{} & \multicolumn{1}{l}{}       & \multicolumn{1}{l}{}                              & \multicolumn{1}{l}{}       & \multicolumn{1}{l}{}                          \\ \hline
\multicolumn{1}{|c|}{\multirow{10}{*}{\textbf{\begin{tabular}[c]{@{}c@{}}\texttt{Gemma}\\ (7B)\end{tabular}}}}    & \multicolumn{1}{c|}{\multirow{2}{*}{\begin{tabular}[c]{@{}c@{}}ARC\\ Challenge\end{tabular}}}    & \multicolumn{1}{c|}{\texttt{Success}}                                                      & 66,7                 & 52,7                       & \multicolumn{1}{c|}{59,2}                         & 68,0                       & \multicolumn{1}{c|}{\textbf{71,8**}}          \\
\multicolumn{1}{|c|}{}                                                                                            & \multicolumn{1}{c|}{}                                                                            & \multicolumn{1}{c|}{\texttt{Error}}                                                        & 64,6                 & 52,7                       & \multicolumn{1}{c|}{65,6}                         & 67,7*                      & \multicolumn{1}{c|}{\textbf{71,8***}}         \\ \cline{2-8} 
\multicolumn{1}{|c|}{}                                                                                            & \multicolumn{1}{c|}{\multirow{2}{*}{\begin{tabular}[c]{@{}c@{}}Causal \\ Judgment\end{tabular}}} & \multicolumn{1}{c|}{\texttt{Success}}                                                      & 55,3                 & \textbf{60,5*}             & \multicolumn{1}{c|}{52,6}                         & \textbf{60,5*}             & \multicolumn{1}{c|}{\textbf{60,5*}}           \\
\multicolumn{1}{|c|}{}                                                                                            & \multicolumn{1}{c|}{}                                                                            & \multicolumn{1}{c|}{\texttt{Error}}                                                        & 44,7                 & 55,3                       & \multicolumn{1}{c|}{47,4}                         & \textbf{57,9*}             & \multicolumn{1}{c|}{\textbf{57,9*}}           \\ \cline{2-8} 
\multicolumn{1}{|c|}{}                                                                                            & \multicolumn{1}{c|}{\multirow{2}{*}{CQA}}                                                        & \multicolumn{1}{c|}{\texttt{Success}}                                                      & 54,7                 & 51,7                       & \multicolumn{1}{c|}{53,7}                         & 56,3                       & \multicolumn{1}{c|}{\textbf{61,0*}}           \\
\multicolumn{1}{|c|}{}                                                                                            & \multicolumn{1}{c|}{}                                                                            & \multicolumn{1}{c|}{\texttt{Error}}                                                        & 54,0                 & 48,3                       & \multicolumn{1}{c|}{57,7*}                        & 57,7*                      & \multicolumn{1}{c|}{\textbf{65,0***}}         \\ \cline{2-8} 
\multicolumn{1}{|c|}{}                                                                                            & \multicolumn{1}{c|}{\multirow{2}{*}{SIQA}}                                                       & \multicolumn{1}{c|}{\texttt{Success}}                                                      & 61,7                 & 67,7**                     & \multicolumn{1}{c|}{63,3}                         & 62,0                       & \multicolumn{1}{c|}{\textbf{68,7***}}         \\
\multicolumn{1}{|c|}{}                                                                                            & \multicolumn{1}{c|}{}                                                                            & \multicolumn{1}{c|}{\texttt{Error}}                                                        & 55,3                 & \textbf{62,7**}            & \multicolumn{1}{c|}{58,0}                         & 56,7                       & \multicolumn{1}{c|}{58,7}                     \\ \cline{2-8} 
\multicolumn{1}{|c|}{}                                                                                            & \multicolumn{1}{c|}{\multirow{2}{*}{Snarks}}                                                     & \multicolumn{1}{c|}{\texttt{Success}}                                                      & 36,1                 & \textbf{50,0***}           & \multicolumn{1}{c|}{38,9}                         & 38,9                       & \multicolumn{1}{c|}{44,4**}                   \\
\multicolumn{1}{|c|}{}                                                                                            & \multicolumn{1}{c|}{}                                                                            & \multicolumn{1}{c|}{\texttt{Error}}                                                        & 36,1                 & 36,1                       & \multicolumn{1}{c|}{41,7}                         & 44,4**                     & \multicolumn{1}{c|}{\textbf{50,0**}}          \\ \hline
\multicolumn{1}{l}{}                                                                                              & \multicolumn{1}{l}{}                                                                             & \multicolumn{1}{l}{}                                                                       & \multicolumn{1}{l}{} & \multicolumn{1}{l}{}       & \multicolumn{1}{l}{}                              & \multicolumn{1}{l}{}       & \multicolumn{1}{l}{}                          \\ \hline
\multicolumn{1}{|c|}{\multirow{10}{*}{\textbf{\begin{tabular}[c]{@{}c@{}}\texttt{Gemma}\\ (2B)\end{tabular}}}}    & \multicolumn{1}{l|}{\multirow{2}{*}{\begin{tabular}[c]{@{}l@{}}ARC\\ Challenge\end{tabular}}}    & \multicolumn{1}{c|}{\texttt{Success}}                                                      & \textbf{41,8}        & 36,1                       & \multicolumn{1}{c|}{32,7}                         & 37,4                       & \multicolumn{1}{c|}{37,1}                     \\
\multicolumn{1}{|c|}{}                                                                                            & \multicolumn{1}{l|}{}                                                                            & \multicolumn{1}{c|}{\texttt{Error}}                                                        & 38,4                 & \textbf{38,8}              & \multicolumn{1}{c|}{34,7}                         & 34,7                       & \multicolumn{1}{c|}{36,1}                     \\ \cline{2-8} 
\multicolumn{1}{|c|}{}                                                                                            & \multicolumn{1}{l|}{\multirow{2}{*}{\begin{tabular}[c]{@{}l@{}}Causal\\ Judgment\end{tabular}}}  & \multicolumn{1}{c|}{\texttt{Success}}                                                      & 42,1                 & 42,1                       & \multicolumn{1}{c|}{47,4}                         & 44,7                       & \multicolumn{1}{c|}{\textbf{52,6*}}           \\
\multicolumn{1}{|c|}{}                                                                                            & \multicolumn{1}{l|}{}                                                                            & \multicolumn{1}{c|}{\texttt{Error}}                                                        & 36,8                 & 34,2                       & \multicolumn{1}{c|}{\textbf{55,3**}}              & 44,7*                      & \multicolumn{1}{c|}{52,6**}                   \\ \cline{2-8} 
\multicolumn{1}{|c|}{}                                                                                            & \multicolumn{1}{l|}{\multirow{2}{*}{CQA}}                                                        & \multicolumn{1}{c|}{\texttt{Success}}                                                      & 39,7                 & \textbf{44,3**}            & \multicolumn{1}{c|}{23,7}                         & 26,0                       & \multicolumn{1}{c|}{41,3}                     \\
\multicolumn{1}{|c|}{}                                                                                            & \multicolumn{1}{l|}{}                                                                            & \multicolumn{1}{c|}{\texttt{Error}}                                                        & \textbf{39,7}        & 37,0                       & \multicolumn{1}{c|}{31,3}                         & 33,0                       & \multicolumn{1}{c|}{32,7}                     \\ \cline{2-8} 
\multicolumn{1}{|c|}{}                                                                                            & \multicolumn{1}{l|}{\multirow{2}{*}{SIQA}}                                                       & \multicolumn{1}{c|}{\texttt{Success}}                                                      & \textbf{59,3}        & 50,0                       & \multicolumn{1}{c|}{50,7}                         & 49,0                       & \multicolumn{1}{c|}{53,0}                     \\
\multicolumn{1}{|c|}{}                                                                                            & \multicolumn{1}{l|}{}                                                                            & \multicolumn{1}{c|}{\texttt{Error}}                                                        & \textbf{52,7}        & 50,7                       & \multicolumn{1}{c|}{51,3}                         & 49,3                       & \multicolumn{1}{c|}{49,7}                     \\ \cline{2-8} 
\multicolumn{1}{|c|}{}                                                                                            & \multicolumn{1}{l|}{\multirow{2}{*}{Snarks}}                                                     & \multicolumn{1}{c|}{\texttt{Success}}                                                      & 27,8                 & 25,0                       & \multicolumn{1}{c|}{\textbf{36,1}}                & \textbf{36,1}              & \multicolumn{1}{c|}{\textbf{36,1}}            \\
\multicolumn{1}{|c|}{}                                                                                            & \multicolumn{1}{l|}{}                                                                            & \multicolumn{1}{c|}{\texttt{Error}}                                                        & 27,8                 & 33,3                       & \multicolumn{1}{c|}{\textbf{38,9}}                & 36,1                       & \multicolumn{1}{c|}{25,0}                     \\ \hline
\end{tabular}
\vspace*{0.5em} 
\caption{\label{tab:results_all} 
\method\ and competitors accuracy (\%) on five test sets, on \texttt{Mistral-7B}, \texttt{Zephyr-7B}, \texttt{Gemma-7B} and \texttt{Gemma-2B}. \method\ is tested on 2 versions, depending on the post hoc explainer used to generate rationales. \texttt{IO} stands for "input-output" standard prompting. \texttt{Auto-CoT} and \texttt{AMPLIFY} are two competing methods automatically generating rationales to enhance the input prompt. The best results are highlighted in bold. With $p$ as the $p$-value of the one-tailed paired $t$-test, *$p<10$\%, **$p<5$\%, ***$p<1$\%. \texttt{IO (ref.)} stands for the reference baseline.}
\end{table*}

\begin{figure*}[h]
\centering
\includegraphics[width=0.9\linewidth]{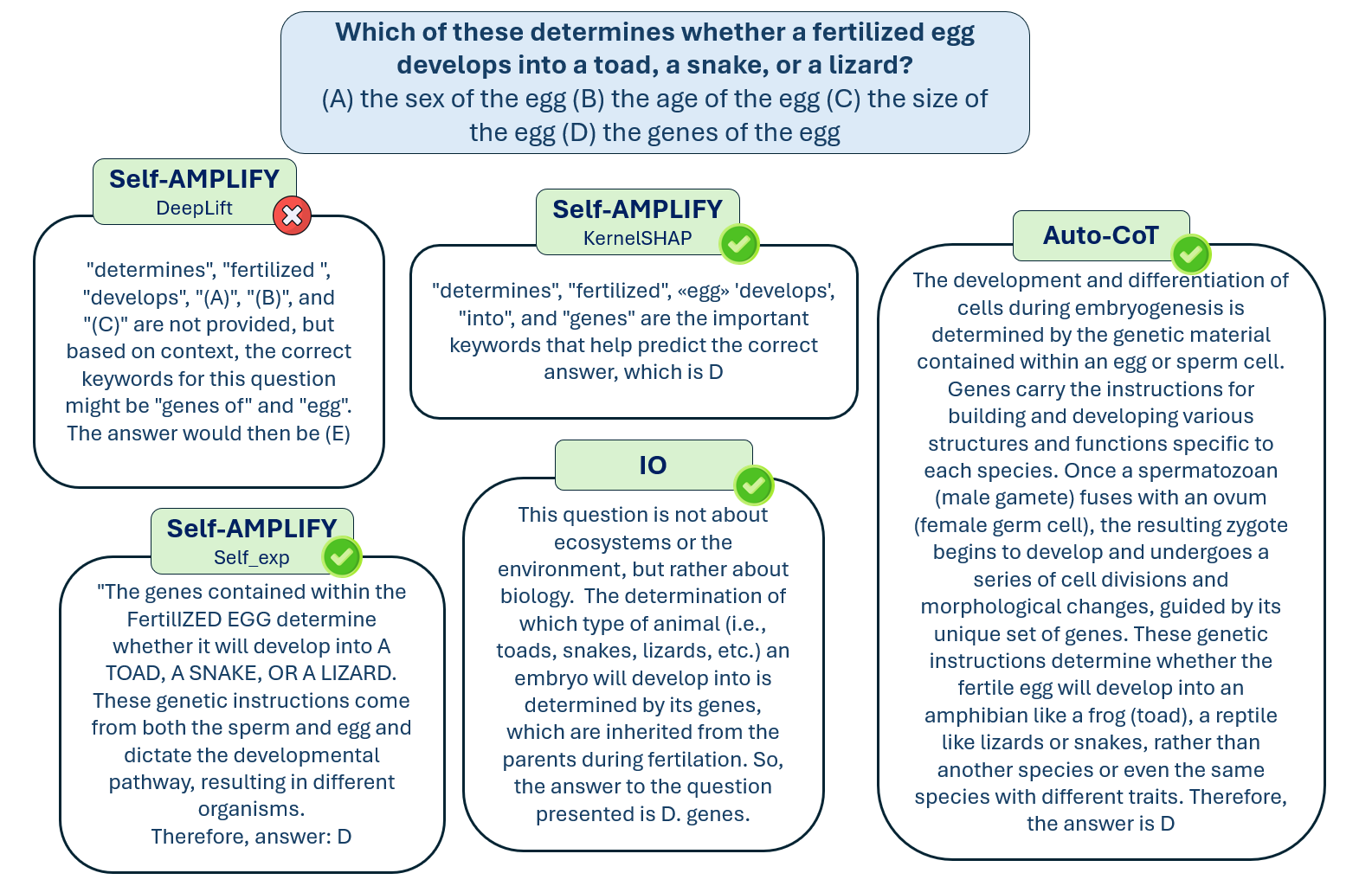}
\caption{ARC Challenge answers conditioned by different ICL prompt built from different rationale generators.}
\label{fig:text_example_arc}
\end{figure*}

\begin{figure*}[h]
\centering
\includegraphics[width=0.9\linewidth]{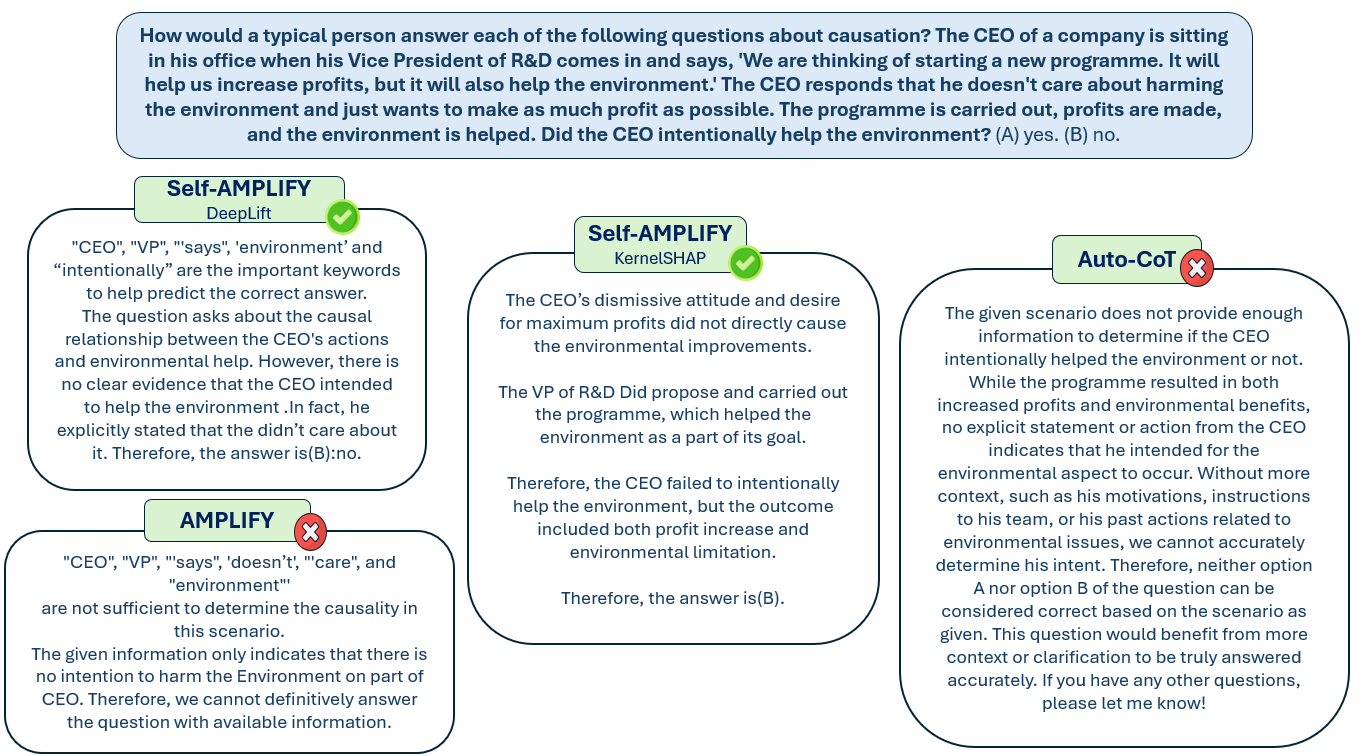}
\caption{Causal Judgment answers conditioned by different ICL prompt built from different rationale generators.}
\label{fig:text_example_causal_judg}
\end{figure*}

\begin{figure*}[t]
\centering
\includegraphics[width=0.9\linewidth]{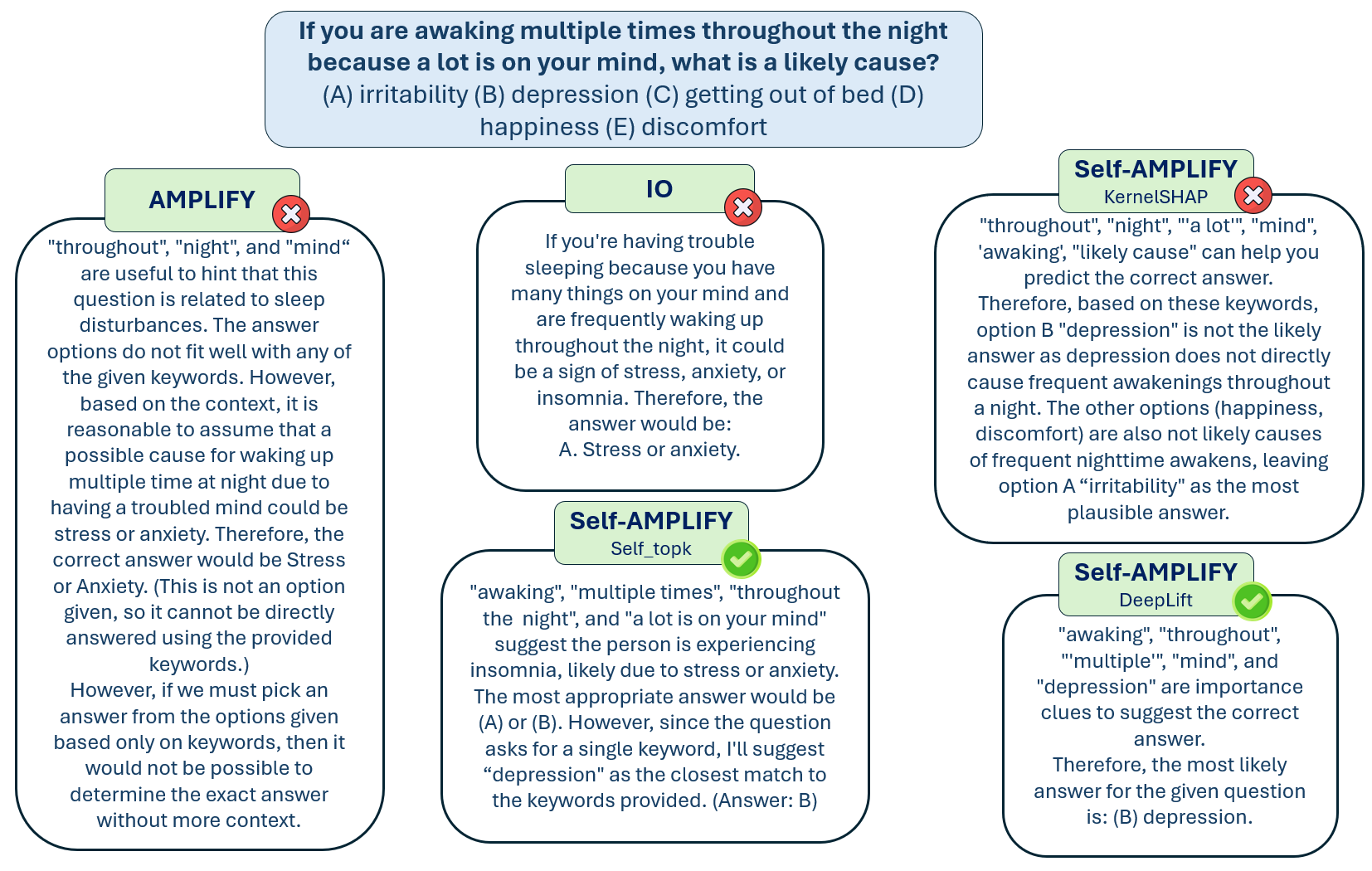}
\caption{Commonsense QA answers conditioned by different ICL prompt built from different rationale generators.}
\label{fig:text_example_cqa}
\end{figure*}

\begin{figure*}[t]
\centering
\includegraphics[width=0.9\linewidth]{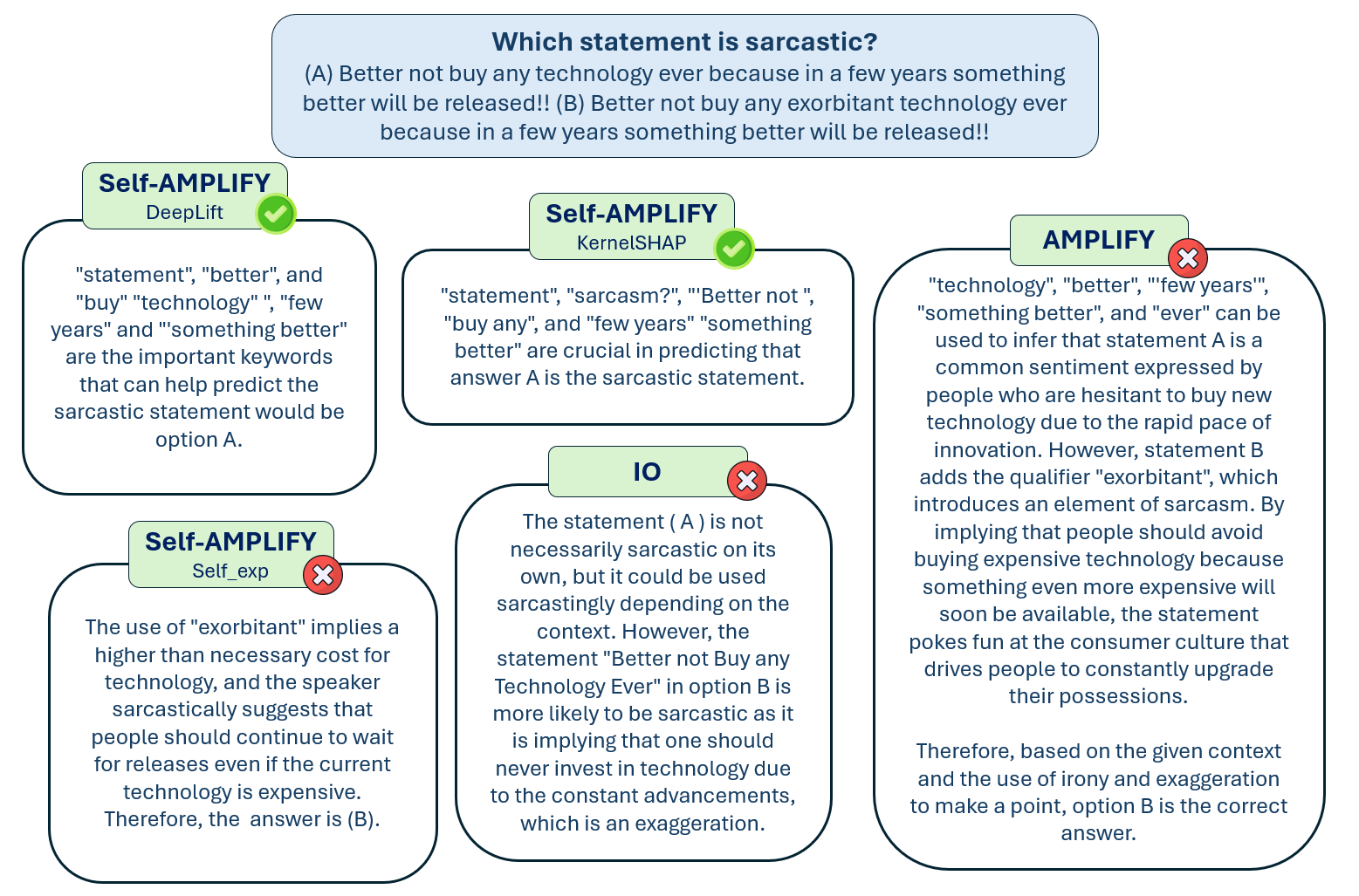}
\caption{Snarks answers conditioned by different ICL prompt built from different rationale generators.}
\label{fig:text_example_snarks}
\end{figure*}

\begin{figure*}[t]
\centering
\includegraphics[width=0.9\linewidth]{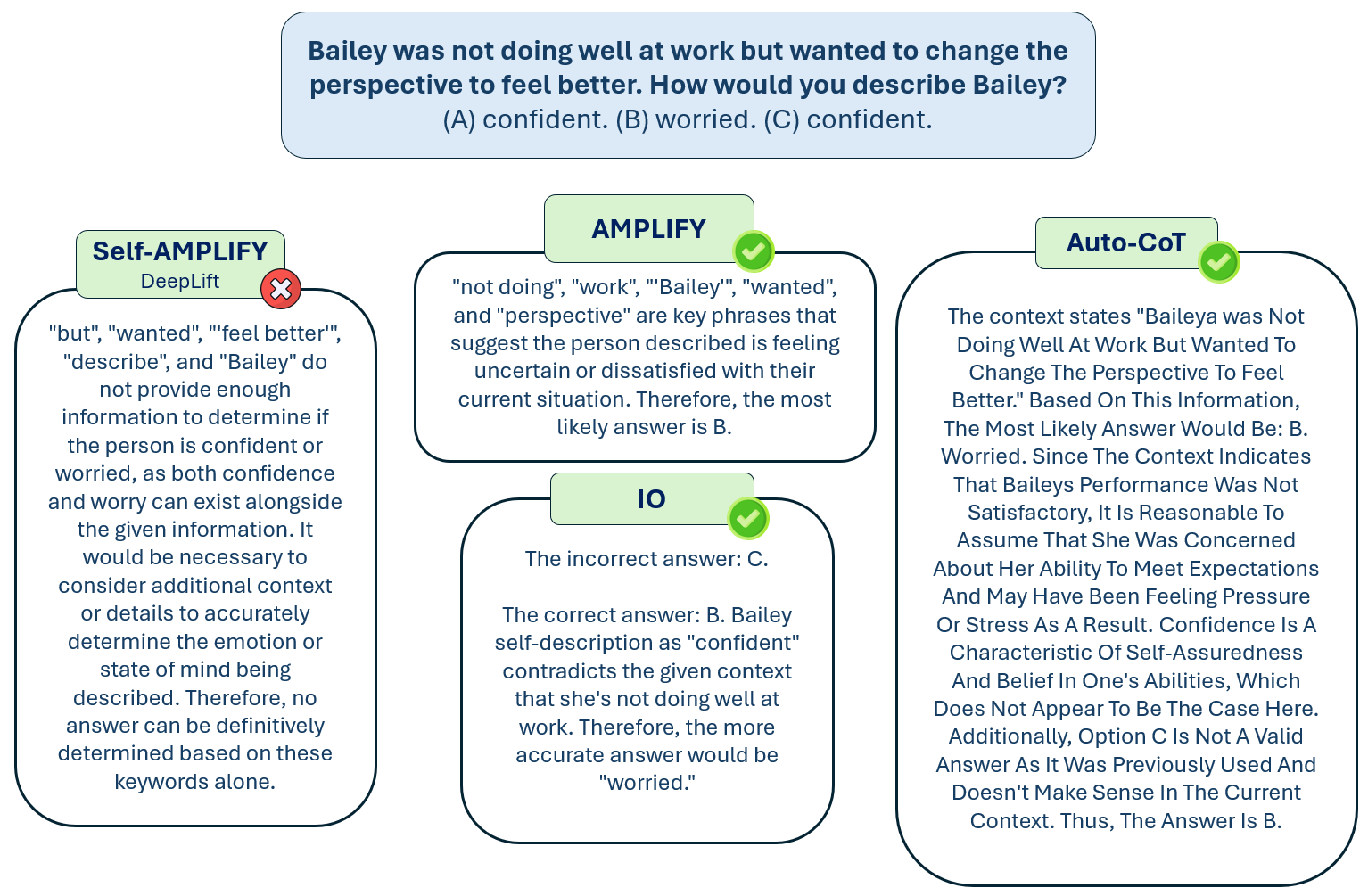}
\caption{SIQA answers conditioned by different ICL prompt built from different rationale generators.}
\label{fig:text_example_siqa}
\end{figure*}

\end{document}